\documentclass[preprint,5p,twocolumn]{elsarticle}

\usepackage{url}
\usepackage{xcolor}
\usepackage{amssymb}
\usepackage{bm}
\usepackage{amsmath}
\usepackage{makecell}
\usepackage{subfig}
\usepackage{float}
\usepackage[utf8]{inputenc}
\usepackage{graphicx}
\usepackage{epstopdf}
\usepackage{stfloats}
\usepackage{multirow}
\usepackage{caption}
\usepackage{stfloats}
\usepackage{booktabs}
\usepackage{tabularx}
\usepackage{hyperref}
\hypersetup{hidelinks}
\journal{Computer Vision and Image Understanding}

\begin{document}

\begin{frontmatter}



\title{Tensor Robust PCA with Nonconvex and Nonlocal Regularization}


\author[1,2]{Xiaoyu Geng}

\author[1,2]{Qiang Guo \corref{cor1}}
\ead{guoqiang@sdufe.edu.cn}
\cortext[cor1]{Corresponding author}

\author[1,2]{Shuaixiong Hui}

\author[3]{Ming Yang}

\author[2,4]{Caiming Zhang}

\affiliation[1]{organization={School of Computer Science and Technology, Shandong University of Finance and Economics},
	postcode={250014},
	city={Jinan}, 
	country={China}}

\affiliation[2]{organization={Shandong Provincial Key Laboratory of Digital Media Technology},
	postcode={250014}, 
	city={Jinan},
	country={China}}

\affiliation[3]{organization={College of Mathematical Sciences, Harbin Engineering University},
	postcode={150001}, 
	country={China}}

\affiliation[4]{organization={School of Software, Shandong University},
	postcode={250101},
	city={Jinan}, 
	country={China}}

\begin{abstract}
Tensor robust principal component analysis (TRPCA) is a classical way for low-rank tensor recovery, which minimizes the convex surrogate of tensor rank by shrinking each tensor singular value equally. However, for real-world visual data, large singular values represent more significant information than small singular values. In this paper, we propose a nonconvex TRPCA (N-TRPCA) model based on the tensor adjustable logarithmic norm. Unlike TRPCA, our N-TRPCA can adaptively shrink small singular values more and shrink large singular values less. In addition, TRPCA assumes that the whole data tensor is of low rank. This assumption is hardly satisfied in practice for natural visual data, restricting the capability of TRPCA to recover the edges and texture details from noisy images and videos. To this end, we integrate nonlocal self-similarity into N-TRPCA, and further develop a nonconvex and nonlocal TRPCA (NN-TRPCA) model. Specifically, similar nonlocal patches are grouped as a tensor and then each group tensor is recovered by our N-TRPCA. Since the patches in one group are highly correlated, all group tensors have strong low-rank property, leading to an improvement of recovery performance. Experimental results demonstrate that the proposed NN-TRPCA outperforms existing TRPCA methods in visual data recovery. The demo code is available at \url{https://github.com/qguo2010/NN-TRPCA}.

\end{abstract}



\begin{keyword}
Low-rank property\sep Nonconvex surrogate\sep Nonlocal self-similarity\sep Tensor robust PCA


\end{keyword}

\end{frontmatter}


\section{Introduction}
As an important method for dimensionality reduction, principal component analysis (PCA) has received extensive attention in the fields of image processing and computer vision \cite{re4, re9, re3}. It is a powerful non-parametric tool to analyze the data corrupted with slight noise. Unfortunately, PCA is sensitive to outliers or large amounts of noise which are inevitably introduced into visual data during acquisition and transmission.

To alleviate this issue, robust PCA (RPCA) \cite{re1} was proposed to recover a low-rank matrix from its observation corrupted by sparse noise, in which the definition of matrix rank is unique. Since the matrix rank function is difficult to be minimized, RPCA adopts nuclear norm as its convex surrogate. Suppose that an observation matrix $\bm{X}\in \mathbb{R}^{{n_1} \times{n_2}}$ can be decomposed by $\bm{X}=\bm{L}+\bm{E} $, where $\bm{L}$ is a low-rank matrix and $\bm{E}$ is a sparse matrix (noise), RPCA obtains $\bm{L}$ and $\bm{E}$ by solving the following minimization objective:
\begin{equation}
	\min_{\bm{L},\bm{E}} \|\bm{L}\Vert_{\ast}+ \lambda\|\bm{E}\Vert_1,\, s.t.\,\bm{X}=\bm{L}+\bm{E},
	\label{eq1}
\end{equation}
where $\lambda$ is a regularization parameter, $\|\bm{L}\Vert_{\ast}$ and $\|\bm{E}\Vert_1$ indicate the matrix nuclear norm of $\bm{L}$ and the $\ell_{1}$-norm of $\bm{E}$, respectively. Problem (\ref{eq1}) can be solved by the SVT algorithm presented in \cite{re33}. Till now, RPCA and its extensions have plenty of applications, including image restoration/alignment \cite{re7}, background foreground separation \cite{re79}, and subspace clustering \cite{re2}. Nevertheless, RPCA can only deal with two-dimensional data. In real-world applications, high-dimensional data is growing explosively. Instead of matrix, tensor is the most appropriate representation of high-dimensional data. For instance, gray videos are three-dimensional tensors with row, column, and temporal modes, while color images are also three-dimensional tensors with size of $height\times width\times channel$. To handle these tensor data, one can apply the RPCA method on each frontal slice of tensors independently. But such a strategy will ignore the multidimensional structure information underlying the tensors. Therefore, it is natural to extend the RPCA from matrix domain to tensor domain.

Given an observed tensor  $\bm{\mathcal{X}}\in \mathbb{R}^{{n_1}\times{n_2}\times{n_3}}$ that is a combination of a low-rank tensor $\bm{\mathcal{L}}$ and a sparse tensor $\bm{\mathcal{E}}$, i.e., $\bm{\mathcal{X}}=\bm{\mathcal{L}}+\bm{\mathcal{E}}$, tensor robust PCA (TRPCA) aims to estimate $\bm{\mathcal{L}}$ and $\bm{\mathcal{E}}$ from tensor $\bm{\mathcal{X}}$. Unlike the matrix rank being unique, tensor rank has many definitions that are derived from different tensor decomposition methods. Tucker rank \cite{re13} is induced for Tucker decomposition \cite{re52}, which is defined as a vector of the matrix rank unfolded along each mode of the given tensor. As minimizing Tucker rank is NP-hard, the sum of nuclear norm (SNN) \cite{re12} was presented as a relaxation of the Tucker rank. Based on SNN, \cite{re11} built a SNN-TRPCA model as follows,
\begin{equation}
	\min_{\bm{\mathcal{L}},\bm{\mathcal{E}}}\sum_{i = 1}^{n_3}\lambda _i\| {\bm{L}^{\{ i \}}}\Vert _* + \| \bm{\mathcal{E}}\Vert_1,\; s.t.\; \bm{\mathcal{X}}=\bm{\mathcal{L}}+\bm{\mathcal{E}},
	\label{eq2}
\end{equation}
where $\lambda_i>0$ and $\bm{L}^{\{ i \}}$ indicates the mode-$i$ matricization \cite{re13} of tensor $\bm{\mathcal{L}}$. SNN-TRPCA exploits the low-rankness of tensor along each dimension. However, it is hard to set the weights $\lambda_i$ due to the fact that the low-rankness of each dimension is usually different in real data. For example, the rank of a gray video along its temporal dimension is much lower than those along its spatial dimensions. Besides, the unfolding operation along one dimension could destroy the inherent structure information of tensors.

Recently, the tensor average rank \cite{re5} was defined by t-SVD \cite{re48}, in which frontal slices of the tensor are arranged in a circulant way by using the block circulant matricization. As a result, this rank can preserve more structural information across frontal slices compared with Tucker rank. Since it is NP-hard to minimize the tensor average rank, the tensor nuclear norm (TNN) \cite{re5} is adopted as a surrogate of the tensor average rank. Based on TNN, a TNN-TRPCA model is introduced as follows:
\begin{equation}
	\min_{{\bm{\mathcal{L}}},{\bm{\mathcal{E}}}} \|\bm{\mathcal{L}}\Vert_T+ \lambda\|\bm{\mathcal{E}}\Vert_1,\; s.t.\;\bm{\mathcal{X}}=\bm{\mathcal{L}}+\bm{\mathcal{E}},
	\label{eq3}
\end{equation}
where $\|\bm{\mathcal{L}} \Vert_T$ denotes the TNN of $\bm{\mathcal{L}}$ (See Definition 5 for details). Problem (\ref{eq3}) can be efficiently solved by the alternating direction method of multipliers (ADMM)\cite{re15}, in which the t-SVT \cite{re5} is a key step. Mathematically,  let $\bm{\mathcal{Y}}=\bm{\mathcal{U}}*\bm{\mathcal{S}}*\bm{\mathcal{V}}^\top$ be the t-SVD of $\bm{\mathcal{Y}}\in{\mathbb{R}^{{n_1}\times{n_2}\times{n_3}}}$, for any $\tau>0$, t-SVT is expressed as follows
\begin{equation}
	\mathcal{D}_{\footnotesize{\tau}}(\bm{\mathcal{Y}})=\bm{\mathcal{U}} * \bm{\mathcal{S}}_{\footnotesize{\tau}} * \bm{\mathcal{V}}^\top,
\end{equation}
where
\begin{equation}
	\bm{\mathcal{S}}_{\footnotesize{\tau }}=\text{ifft}\big((\bar{\bm{\mathcal{S}}}-\tau)_+, [\,], 3\big),
	\label{eq15}
\end{equation}
$\bar{\bm{\mathcal{S}}}$ is the result of fast Fourier transform (FFT) on $\bm{\mathcal{S}}$ along the 3-rd dimension, and ifft is the inverse operator of FFT. It is easy to see that t-SVT shrinks each singular value equally according to the threshold $\tau$. However, in practice, the tensor singular values often have different physical meanings. For a noisy color image, the large singular values usually correspond to the important information in the image, while the small singular values usually represent the noise.  This motivates us to utilize different thresholds to shrink the small singular values more and the large ones less, so that the noise can be reduced precisely and the siginificant information can be preserved well. Besides, TNN-TRPCA has a potential limitation that it simply assumes the whole underlying tensor is of low rank. For visual data (e.g., natural images and videos), such an assumption is often difficult to be satisfied. Therefore, TNN-TRPCA cannot well recover the detail information in visual data, especially in data with complex stuctures.

For solving the above problems, we intend to propose a variant of TRPCA by shrinking tensor singular values differently and integrating nonlocal self-similarity. Specifically, to better preserve the important information of tensor data, a nonconvex TRPCA (N-TRPCA) model is built using tensor adjustable logarithmic norm as a nonconvex surrogate of the tensor average rank. It can apply adaptive thresholds for shrinking different tensor singular values. Then, the nonlocal self-similarity is further introduced into N-TRPCA to derive a nonconvex and nonlocal TRPCA (NN-TRPCA) model. By this way, our model can make full use of the structural redundancy of tensors to recover the detail information, resulting in remarkable performance improvements. In summary, our contributions are highlighted as follows:

\begin{itemize}
	\item A nonconvex TRPCA (N-TRPCA) model under t-SVD framework is proposed for visual data recovery, which makes the large singular values shrink less and the small singular values shrink more simultaneously. Such a model can effectively preserve the important information in visual data.
	
	\item Beyond using the global low-rankness of tensors, nonlocal low-rank property is more crucial to fully utilize the structural redundancy in tensors. Thus, we further incorporate the nonlocal self-similarity into the N-TRPCA and then propose a nonconvex
	and nonlocal TRPCA model, named NN-TRPCA. 
	
	\item To solve the proposed NN-TRPCA, we present an effective ADMM-based algorithm, in which the variables have the closed-form equations.
	
	\item We evaluate the efficacy of the N-TRPCA and NN-TRPCA in color image restoration and gray video restoration. Extensive experiment results confirm the superiority of our methods and show their competitive performance with the state-of-the-arts.
\end{itemize}

A preliminary conference version of this work was presented in \cite{re28}. We extend it both theoretically and experimentally. First, the recent works on the low-rank prior and the nonlocal prior of visual data are elaborated to provide theoretical basis for the proposed methods. Second, we offer a detailed and rigorous derivation for the closed-form solutions of our N-TRPCA algorithm. Third, we compare two methods of constructing group tensors and offer a new observation. Fourth, the proposed N-TRPCA and NN-TRPCA are evaluated in visual data restoration, where the data is corrupted with random noise. This restoration task is more challenging than denoising visual data corrupted by Gaussian white noise. Fifth, we analysis the influence of parameters in NN-TRPCA on our experiments. Finally, we discuss several possible future extensions of our NN-TRPCA.

The remainder of this paper is organized as follows. Some notations and preliminaries are given in Section \ref{sec2}. The recent works on the  low-rankness and the nonlocal self-similarity are reviewed in Section \ref{sec3}. In Section \ref{sec4}, we propose the N-TRPCA and NN-TRPCA methods and design the corresponding optimization algorithms. Section \ref{sec5} reports extensive experimental results. Finally, Section \ref{sec6} draws a conclusion.

\section{Preliminaries}\label{sec2}

For convenience of presentation, we first introduce some notations used in our work, and then list some basic definitions and theorems of the tensor algebra.

Third-order tensors are denoted as boldface calligraphic letters, e.g., $\bm{\mathcal{A}}\in{\mathbb{R}^{{n_1}\times{n_2}\times{n_3}}}$, matrices are as boldface capital letters, e.g., $\bm{A}\in{\mathbb{R}^{{n_1}\times{n_2}}}$, vectors are as boldface lowercase letters, e.g., $\bm{a}\in{\mathbb{R}^{n_1}}$, and scalars are as lowercase letters, e.g., $a\in{\mathbb{R}}$. 
For a three-order tensor $\bm{\mathcal{A}}$, we represent $\bar{\bm{\mathcal{A}}}$ as the Fast Fourier Transform (FFT) of $\bm{\mathcal{A}}$ along 3-rd dimension using the Matlab command fft, i.e., $\bar{\bm{\mathcal{A}}}=\text{fft}(\bm{\mathcal{A}}, [\;], 3)$, and achieve $\bm{\mathcal{A}}$ by the inverse FFT, i.e., $\bm{\mathcal{A}}=\text{ifft}(\bar{\bm{\mathcal{A}}}, [\;], 3)$. $\bm{A}^{(i)}$ and $\bar{\bm{A}}^{(i)}$ are respectively  the $i$th frontal slice of $\bm{\mathcal{A}}$ and $\bar{\bm{\mathcal{A}}}$. Moreover, the inner product between $\bm{\mathcal{A}}$ and $\bm{\mathcal{B}}$ is represented as $<\bm{\mathcal{A}},\bm{\mathcal{B}}>=\sum_{i=1}^{n_3}<\bm{A}^{(i)},\bm{B}^{(i)}>$. The $\ell_{1}$-norm, infinity norm, and Frobenius norm of $\bm{\mathcal{A}}$ are defined as $\|\bm{\mathcal{A}}\Vert_1=\sum_{ijk}|a_{ijk}|$, $\|\bm{\mathcal{A}}\Vert_{\infty}=\max_{ijk}|a_{ijk}|$, and $\|\bm{\mathcal{A}}\Vert_ F=\sqrt{\sum_{ijk}|a_{ijk}|^2}$,  respectively, where $a_{ijk}$ denotes the $(i,j,k)$th entry of $\bm{\mathcal{A}}$. For a matrix $\bm{A}$, its nuclear norm is defined as the sum of singular values, i.e.,  $\|\bm{A}\|_{*}=\sum_{i} \sigma_{i}(\bm{A})$, where $\sigma_j(\cdot)$ is the $i$th largest singular value of  $\bm{A}$.

\noindent\textbf{Definition 1. (Block diagonal matrix \cite{re5})}
For $\bm{\mathcal{A}}\in{\mathbb{R}^{{n_1}\times{n_2}\times{n_3}}}$, its block diagonal matrix is defined as
\begin{equation}
	\bar{\bm{A}}=\text{bdiag}(\bar{\bm{\mathcal{A}}})=\left[\begin{array}{llll}
		\bar{\bm{A}}^{(1)} & & & \\
		& \bar{\bm{A}}^{(2)} & & \\
		& & \ddots & \\
		& & & \bar{\bm{A}}^{(n_{3})}
	\end{array}\right]\in \mathbb{C}^{n_{1} n_{3} \times n_{2} n_{3}}.
\end{equation}

\noindent\textbf{Definition 2. (Block circulant matrix \cite{re5})} For $\bm{\mathcal{A}}\in{\mathbb{R}^{{n_1}\times{n_2}\times{n_3}}}$, the 
block circulant matrix of $\bm{\mathcal{A}}$ is 
\begin{equation}
	\text{bcirc}(\bm{\mathcal{A}})=\left[\begin{array}{cccc}
		\bm{A}^{(1)} & \bm{A}^{(n_{3})} & \cdots & \bm{A}^{(2)} \\
		\bm{A}^{(2)} & \bm{A}^{(1)} & \cdots & \bm{A}^{(3)} \\
		\vdots & \vdots & \ddots & \vdots \\
		\bm{A}^{(n_{3})} & \bm{A}^{(n_{3}-1)} & \cdots & \bm{A}^{(1)}
	\end{array}\right]  \in \mathbb{R}^{n_{1} n_{3} \times n_{2} n_{3}}.
\end{equation}

\noindent\textbf{Theorem 1. (Diagonalization \cite{re5})} The block circulant matrix of $\bm{\mathcal{A}}$ can be block diagonalized by the following equation:
\begin{equation}
	(\bm{F}_{n_{3}} \otimes \bm{I}_{n_{1}}) \cdot \text{bcirc}(\bm{\mathcal{A}}) \cdot(\bm{F}_{n_{3}}^{-1} \otimes \bm{I}_{n_{2}})=\bar{\bm{A}},
\end{equation}
where $\bm{F}_{n_{3}}\in \mathbb{C}^{n_{3} \times n_{3}}$ is the discrete fourier transformation matrix, $\bm{I}_{n_{1}}\in \mathbb{R}^{n_{1} \times n_{1} \times n_{3}}$ and $\bm{I}_{n_{2}}\in \mathbb{R}^{n_{2} \times n_{2}\times n_{3}}$ are two identity matrices, $\otimes$ is the Kronecker product.

\noindent\textbf{Definition 3. (T-product \cite{re48})} Given $\bm{\mathcal{A}}\in{\mathbb{R}^{{n_1}\times{n_2}\times{n_3}}}$ and
$\bm{\mathcal{B}}\in{\mathbb{R}^{{n_2}\times{l}\times{n_3}}}$, the t-product $\bm{\mathcal{A}}*
\bm{\mathcal{B}}$ is defined as a tensor with size ${n_1}\times{l}\times{n_3}$, 
\begin{equation}
	\bm{\mathcal{A}}*\bm{\mathcal{B}} =\text{fold}\big(\text{bcirc}(\bm{\mathcal{A}})\cdot \text{unfold}(\bm{\mathcal{B}})\big).
\end{equation}
Using Theorem 1, the t-product can be transformed into the matrix multiplication in the Fourier domain, i.e., $\bm{\mathcal{A}}*\bm{\mathcal{B}}=\bar{\bm{A}}\bar{\bm{B}}$.

\noindent\textbf{Theorem 2. (T-SVD \cite{re48})} For $\bm{\mathcal{A}}\in{\mathbb{R}^{{n_1}\times{n_2}\times{n_3}}}$, the
tensor singular value decomposition (t-SVD) of $\bm{\mathcal{A}}$ is discribed by
\begin{equation}
	\bm{\mathcal{A}}=\bm{\mathcal{U}}*\bm{\mathcal{S}}*\bm{\mathcal{V}}^\top,
\end{equation}
where $\bm{\mathcal{S}}\in{\mathbb{R}^{{n_1}\times{n_2}\times{n_3}}}$ is an f-diagonal tensor,  $\bm{\mathcal{U}}\in{\mathbb{R}^{{n_1}\times{n_1}\times{n_3}}}$ and  $\bm{\mathcal{V}}\in{\mathbb{R}^{{n_2}\times{n_2}\times{n_3}}}$ are two orthgonal tensors. 

\noindent\textbf{Definition 4. (Tensor average rank \cite{re5})} Given a tensor  $\bm{\mathcal{A}}\in{\mathbb{R}^{{n_1}\times{n_2}\times{n_3}}}$, the tensor average rank of $\bm{\mathcal{A}}$ is as:
\begin{equation}
	\text{rank}_{a}(\bm{\mathcal{A}})=\frac{1}{n_{3}} \text{rank}\big(\text{bcirc}(\bm{\mathcal{A}})\big).
\end{equation}

\noindent\textbf{Definition 5. (Tensor Nuclear Norm (TNN) \cite{re5})} Given a tensor  $\bm{\mathcal{A}}\in{\mathbb{R}^{{n_1}\times{n_2}\times{n_3}}}$, the tensor nuclear norm
of $\bm{\mathcal{A}}$ is depicted as
\begin{align}
	\|\bm{\mathcal{A}}\|_T=&\frac{1}{n_3} \|  \text{bcirc}(\bm{\mathcal{A}})\Vert_* =\frac{1}{n_3}\sum^{n_3}_{i=1} \|\bar{\bm{A}}^{(i)}\Vert_*
	\nonumber\\=&\frac{1}{n_3}\sum^{n_3}_{i=1}\sum^{\min(n_1,n_2)}_{j=1}\sigma_j(\bar{\bm{A}}^{(i)}),
\end{align}
which is a convex surrogate of the tensor average rank.

\section{Related Works}\label{sec3}

The most important issue of noise reduction from corrupted visual data is to fully use the underlying structure priors of the data. Both low-rank prior \cite{ re67,re53, re63, re62} and nonlocal prior \cite{re46, re68, re43, re44} are two commonly-used structure priors for visual data recovery. In the following, we briefly review some related works.

\subsection{Low-Rank Property}\label{sec3.1}

The low-rank prior indicates that the images have some structural redundancy, i.e., repeating structures. Numerous works \cite{re63, re66, re62} have employed this prior for the task of image recovery. Generally, they transform the image recovery into a matrix rank minimization problem. For instance, a gray image can be represented as a low-rank matrix. Similarly, a color image can be approximated by a low-rank matrix on RGB channels independently. Although the matrix rank is capable to characterize the global information in matrices, it is NP-hard to be minimized. Thus, many researchers have attempted to find an appropriate surrogate of the matrix rank. \cite{re29} originally used the matrix nuclear norm (MNN) as a convex surrogate of the matrix rank. In \cite{re30}, it is theoretically proved that MNN is the best convex approximation of the matrix rank. Due to its convexity, the MNN minimization can be efficiently solved by the SVT algorithm \cite{re33} and has a global optimal solution. However, MNN ignores the difference between singular values. For visual data, the large singular values contain more important information than the small singular values. Thus, several variants of MNN were developed to treat the singular values in different manners. Matrix truncated nuclear norm (MTNN) \cite{re34} shrinks only the smallest singular values. Although MTNN achieves better performance than MNN, it ignores the fact that the large singular values contain a small amount of noise. To adaptively shrinks the singular values, \cite{re20} proposed a matrix weighted nuclear norm (MWNN) and derived a weighted SVT (WSVT) algorithm for effectively minimizing the MWNN. By setting the weights to decrease as the singular value increases, the large singular values can be shrunk less, while the small singular values can be shrunk more. But it is troublesome to estimate a set of reasonable weights. Recently, by using logarithm function, a nonconvex approximation of the matrix rank \cite{re77} was proposed to adaptively estimate these weights. Since the estimated weights decrease as the singular values increase, this nonconvex surrogate can simultaneously increase the shrinkage on small singular values and reduce the shrinkage on large singular values.

One shortcoming of the matrix rank based methods is that they cannot preserve the correlations across frontal slices of multidimensional visual data, such as the correlations across RGB channels of color images or the correlations across frames of gray videos. Instead of matrix, tensor is an effective representation form for visual data without loss of its structural information. Hence, approximating visual data directly by low-rank tensors has gained significant popularity in recent years. Nevertheless, the definition of tensor rank is nonunique. The commonly-used definitions are CP rank \cite{re13}, Tucker rank \cite{re13}, and tensor average rank \cite{re5}. CP rank is defined by the smallest number of tensor rank-one decomposition. But predefining CP rank is a challenge. Tucker rank reflects the low-rankness of matrix unfolded along each mode of tensors. Similar to matrix cases, minimizing the Tucker rank is also NP-hard. SNN \cite{re12} was used as a convex approximation of the Tucker rank for low-rank recovery. But SNN fails to capture the intrinsic correlations between different modes. Recently, the tensor average rank \cite{re5} was proposed. Unlike the Tucker rank that adopts matricization along several fixed directions, the tensor average rank can capture more correlations across frontal slices of tensors by block circulant matricization. Since the tensor average rank is NP-hard to be minimized, \cite{re5} proposed TNN as a surrogate of the tensor average rank and further developed a t-SVT algorithm to efficiently solve TNN. However, both SNN and TNN neglect the differences between tensor singular values. As a consequence, they cannot well preserve the important information of tensor data. In order to address this issue, several nonconvex surrogates of tensor rank \cite{re61, re78} were presented to treat the tensor singular values differently.

\subsection{Nonlocal Self-Similarity}\label{sec3.2}

The nonlocal self-similarity of images is another important prior. Compared with global low-rankness, it can capture detailed structural redundancy, resulting in more accurate recovery results. \cite{re41} firstly applied the nonlocal self-similarity to gray image denoising, and presented the NLM filter. Concretely, NLM restores each pixel via nonlocal averaging of the pixels in its neighborhood, where the weights for a pixel reflect the similarity of other pixels with it. Another representative work is BM3D \cite{re42}, which groups similar 2-D image  patches as 3-D tensors and handles these tensors with sparse collaborative filtering. In essence, BM3D combines the sparsity and nonlocal self-similarity of images for image denoising. Futhermore, \cite{re43} proposed NCSR to learn the sparse coding of nonlocal redundancy in the images, and then utilized NCSR to deal with several image restoration tasks. In a recent decade, there has been a growing interest in using both nonlocal self-similarity and low-rankness of matrix in the field of image processing. NLR-CS \cite{re44} introduces the nonlocal self-similarity into compressed sensing recovery. It groups similar 2-D image patches into a matrix and handles the group matrix by low-rank regularization. Due to the similar patches in one group having a strong correlation, each group matrix is strongly low-rank. Similarly, the model presented in \cite{re46} groups similar patches into a matrix, and then estimates each matrix by the low-rank approximation based on singular values truncation.

Recently, nonlocal low-rank matrix recovery has been extended to tensor domain. Unlike the matrix cases, nonlocal low-rank tensor recovery groups similar image patches as a tensor and treats the group tensors as the basic recovery units. For instance, a tensor-based compressed sensing recovery framework (NLR-TFA)  \cite{re68} was proposed by utilizing nonlocal self-similarity and low-CP-rank regularization. Since computing CP rank is NP-hard, NLR-TFA uses Jenrich’s algorithm \cite{re69} to estimate CP rank. Additionally, nonlocal low-rank regularization-based tensor completion (NLRR-TC) \cite{re70} combines nonlocal prior and low-Tucker-rank constraint for hyperspectral image completion, which can simultaneously  capture the spatial and spectral correlations of hyperspectral images. It is noteworthy that the tensor average rank can well describe the correlation across frontal slices of one tensor. \cite{re76} integrated the nonlocal prior into low-rank tensor completion based on the tensor average rank for visual data completion.

In summary, the nonconvex low-rank regularizer and nonlocal self-similarity have been widely studied in the literature. Among them, the most relevant to our approach are the works \cite{re77} and \cite{re76}. Different from the logarithmic function $g(x)=log(x+1)$ using in \cite{re77}, our logarithmic function introduces an adjustable parameter $\theta$ to further control the level of shrinkage on tensor singular values. Besides, although our NN-TRPCA and \cite{re76} are nonlocal tensor recovery models based on the tensor average rank, our NN-TRPCA can better capture the correlations across frontal slices of 3-D visual data, compared to processing each frontal slice separately in \cite{re76}.

\section{Proposed Method}\label{sec4}

\begin{figure*}[!t]
	\centering
	\includegraphics[width=0.78\linewidth]{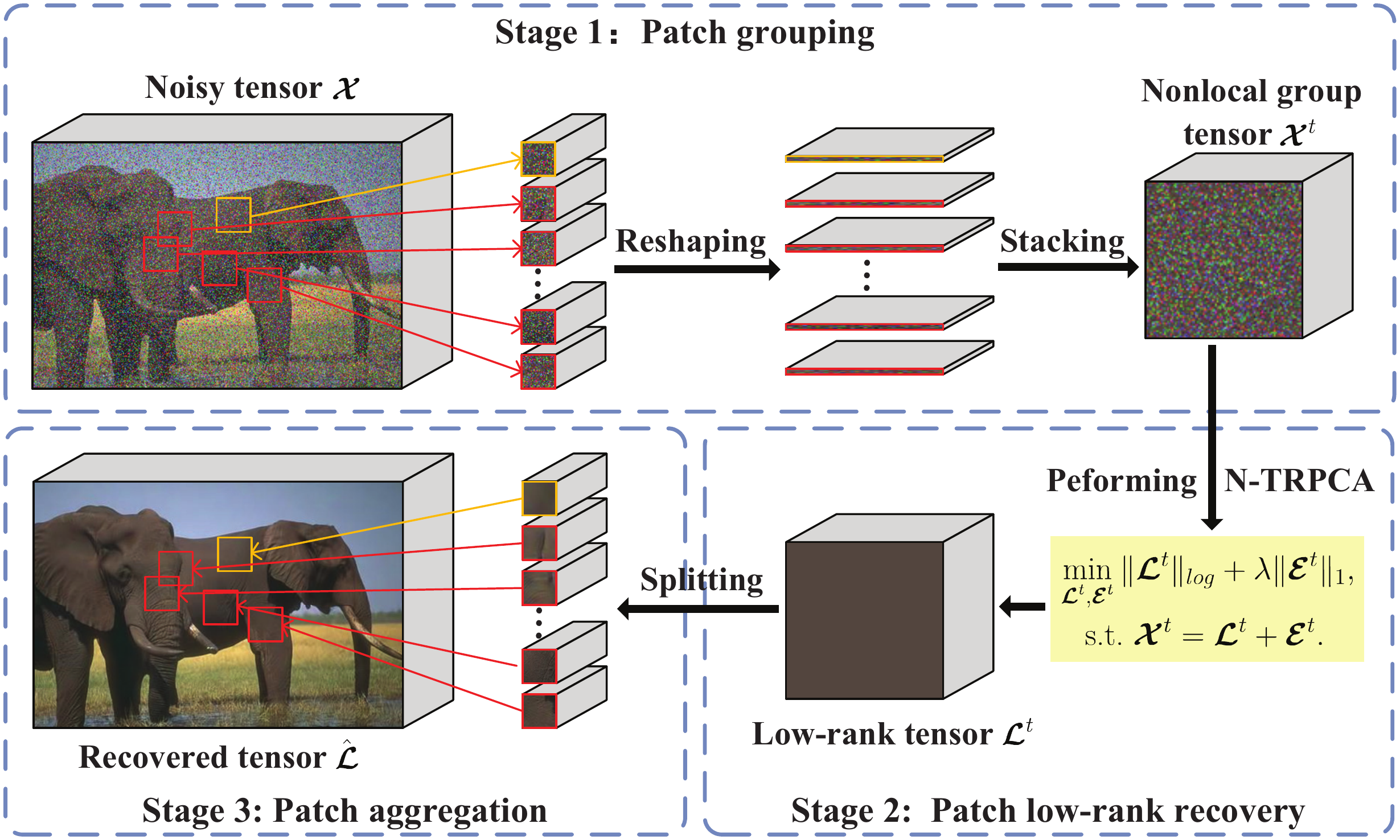}
	\caption{Flowchart of the proposed NN-TRPCA by taking a group of similar patches as an example. The patch with yellow box is a reference patch, and the patches with red boxes are its similar patches.}
	\label{Fig_flowchart}
\end{figure*}

In this section, we first introduce a tensor adjustable logarithmic norm that is a nonconvex surrogate of the tensor average rank and  use it to build a nonconvex TRPCA (N-TRPCA) model. Then an ADMM-based algorithm is developed to efficiently solve our N-TRPCA. Finally, the nonlocal self-similarity is integrated into N-TRPCA to derive the nonconvex and nonlocal TRPCA model, called NN-TRPCA. Fig. \ref{Fig_flowchart} illustrates the flowchart of the proposed NN-TRPCA method.

\subsection{N-TRPCA Model}\label{sec4.1}

By Eq. (\ref{eq15}), we know that t-SVT shrinks each singular value by the same threshold $\tau$ in solving the TNN minimization. But in real scenarios, there exist great differences among tensor singular values. For instance, the large singular values of a noisy image usually deliver significant information, while the small singular values usually correspond the noise. Accordingly, the large singular values should be shrunk less, and the small singular values should be shrunk more. For this goal, we propose the following nonconvex surrogate of the tensor average rank.

\noindent\textbf{Definition 6. (Tensor Adjustable Logarithmic Norm (TALN))} Given any tensor  $\bm{\mathcal{A}}\in{\mathbb{R}^{{n_1}\times{n_2}\times{n_3}}}$, $r=\min(n_1,n_2)$, the tensor adjustable logarithmic norm of $\bm{\mathcal{A}}$ is defined as
\begin{equation}
	\|\bm{\mathcal{A}}\Vert_{log}=\frac{1}{n_3}\sum^{n_3}_{i=1}\sum^{r}_{j=1}g \big(\sigma_j(\bar{\bm{A}}^{(i)})\big),
\end{equation}
where $g(x)=log(\theta x+1)$ is a nonconvex function with adjustable positive parameter $\theta$. 
\label{def7}

\begin{figure}[!tbp]
	\centering
	\includegraphics[width=2.5in]{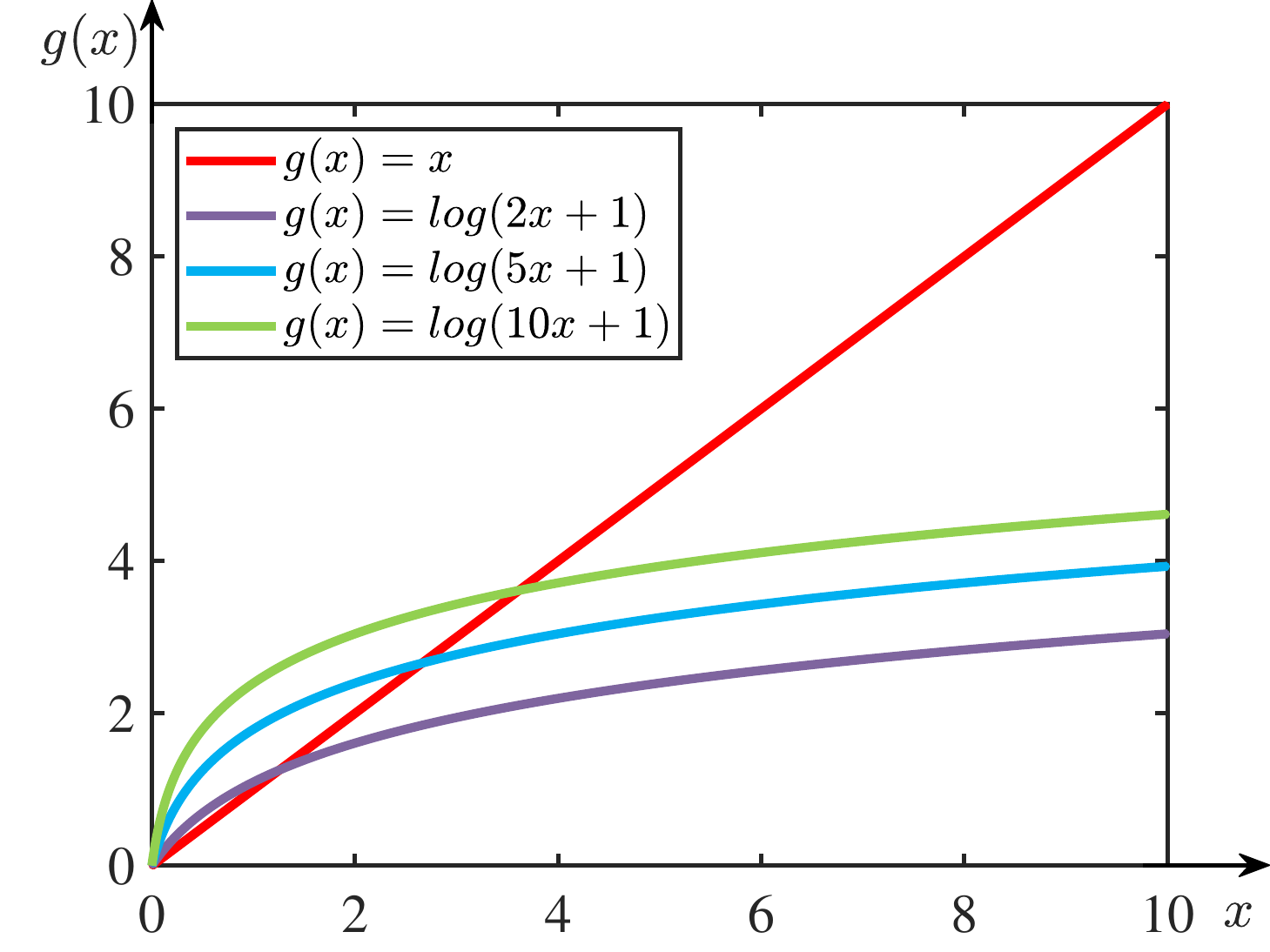}
	\caption{Visual comparison of $g(x)=x$ and the noncovex function $g(x)=log(\theta  x+1)$ with different $\theta$.}
	\label{Fig_log}
\end{figure}

One main advantage of TALN is that it better preserves the important information in tensor data than TNN. In Section \ref{sec4.2}, we will show that TALN can adaptively estimate weight for each singular value,  and further show that the weight decreases as the singular value increases. According to these weights, different thresholds are used for shrinking the tensor singular values, and the smaller thresholds correspond to the larger singular values. As a result, TALN shrinks small singular values more and large singular values less. Another advantage of TALN is that it flexibly controls the shrinkage level of $g(x)$ to tensor singular values by the adjustable parameter $\theta$. Fig. \ref{Fig_log} provides a visual comparison of $g(x)=x$ used in TNN and $g(x)=log(\theta  x+1)$ used in our TALN with different $\theta$. It can be observed that, unlike the behavior of $g(x)=x$, the nonconvex function $g(x)=log(\theta  x+1)$ can increase the shrinkage on small singular values and decrease the shrinkage on large singular values  simultaneously. As $\theta$ increases, $g(x)$ increases the shrinkage on all singular values.

Based on TALN, we propose a N-TRPCA model, i.e., 
\begin{equation}
	\min_{{\bm{\mathcal{L}}},{\bm{\mathcal{E}}}} \|\bm{\mathcal{L}}\Vert_{log}+ \lambda\|\bm{\mathcal{E}}\Vert_1,
	\; s.t.\;\bm{\mathcal{X}}=\bm{\mathcal{L}}+\bm{\mathcal{E}},
	\label{eq4}
\end{equation}
where $\lambda$ is a regularization parameter.

\subsection{Optimization Algorithm of N-TRPCA}\label{sec4.2}

As used in \cite{re80, re84}, ADMM is an efficient approach to solve the optimization problem with multiple constraint terms. In the following, we present an ADMM-based algorithm for solving the model (\ref{eq4}). The augmented Lagrangian function of (\ref{eq4}) is 
\begin{align}
	L(\bm{\mathcal{L}},\bm{\mathcal{E}},\bm{\mathcal{P}},\mu)=&\|\bm{\mathcal{L}}\Vert_{log}+ \lambda\|\bm{\mathcal{E}}\Vert_1+<\bm{\mathcal{P}},\bm{\mathcal{L}}+\bm{\mathcal{E}}-\bm{\mathcal{X}}>\nonumber\\
	&+\frac{\mu}{2}\|\bm{\mathcal{L}}+\bm{\mathcal{E}}-\bm{\mathcal{X}}\Vert^2_F,
	\label{eq5}
\end{align}
where $\bm{\mathcal{P}}$ is a Lagrange multiplier and $\mu$ is a penalty parameter. $\bm{\mathcal{L}}$ and $\bm{\mathcal{E}}$ can be iteratively solved  by minimizing function (\ref{eq5}).

With parameters $\bm{\mathcal{E}}_k$, $\bm{\mathcal{P}}_k$, $\mu_k$  fixed, $\bm{\mathcal{L}}_{k+1}$ can be obtained by
\begin{equation}
	\bm{\mathcal{L}}_{k+1}=\arg\min_{\bm{\mathcal{L}}} \, \|\bm{\mathcal{L}}\Vert_{log}+\frac{\mu_k}{2}\|\bm{\mathcal{L}}-\bm{\mathcal{Q}}_k\Vert^2_F,\\
	\label{eq6}
\end{equation}
where $\bm{\mathcal{Q}}_k=\bm{\mathcal{X}}-\bm{\mathcal{E}}_k-\mu_k^{-1} \bm{\mathcal{P}}_k$, subscript $k$ indicates the $k$th iteration.  Using Definition 6, Eq. (\ref{eq6}) can be rewritten as
\begin{equation}
	\bm{\mathcal{L}}_{k+1}=\arg\min_{\bm{\mathcal{L}}} \, \frac{1}{n_3}\sum_{i=1}^{n_{3}} \sum^{r}_{j=1}g \big(\sigma_j(\bar{\bm{L}}^{(i)})\big)+\frac{\mu_k}{2}\|\bm{\mathcal{L}}-\bm{\mathcal{Q}}_k\Vert^2_F,
	\label{eq7}
\end{equation}
where $r=\min(n_1,n_2)$. For simplicity of description, we denote $\sigma_j^{i}=\sigma_j(\bar{\bm{L}}^{(i)})$ and  $\sigma_{j,k}^{i}=\sigma_{j,k}(\bar{\bm{L}}^{(i)})$. To estimate weights for the singular values $\sigma_j^{i}$, the function $g(\sigma_{j}^{i})$ is approximated by its first-order Taylor expansion, i.e.,
\begin{equation}
	g(\sigma_{j}^{i}) = g(\sigma_{j,k}^{i})+w_{j,k}^{i}(\sigma_{j}^{i}-\sigma_{j,k}^{i}),
	\label{eq8}
\end{equation}
where $	w_{j,k}^{i} = \partial g(\sigma_{j,k}^{i})=\theta /( \theta  \sigma_{j,k}^{i}+1)$ is the derivative at point $\sigma_{j,k}^{i}$ that is equivalent to the weight assigned to $\sigma_{j}^{i}$. Since $\partial g(x)$ is monotonically decreasing and
$\sigma_{1,k}^{i} \geq \sigma_{2,k}^{i} \geq \cdots \geq \sigma_{r,k}^{i} \geq 0$, the weights have the following property
\begin{equation}
	0 \leq  w_{1,k}^{i} \leq w_{2,k}^{i} \leq \cdots \leq w_{r,k}^{i}.
	\label{eq9}
\end{equation} 
Using Eq. (\ref{eq8}), Eq. (\ref{eq7}) is approximated as 
\begin{align}
	\bm{\mathcal{L}}_{k+1}=&\arg\min_{\bm{\mathcal{L}}} \, \frac{1}{n_3}\sum_{i=1}^{n_{3}} \sum^{r}_{j=1}\big(g(\sigma_{j,k}^{i})\nonumber+ w_{j,k}^{i}(\sigma_{j}^{i}-\sigma_{j,k}^{i})\big)
	\\&+\frac{\mu_k}{2}\|\bm{\mathcal{L}}-\bm{\mathcal{Q}}_k\Vert^2_F,\nonumber \\
	=&\arg\min_{\bm{\mathcal{L}}} \, \frac{1}{n_3}\sum_{i=1}^{n_{3}} \sum^{r}_{j=1}w_{j,k}^{i}\sigma_{j}^{i}+\frac{\mu_k}{2}\|\bm{\mathcal{L}}-\bm{\mathcal{Q}}_k\Vert^2_F.
	\label{eq10}
\end{align}
Finally, the minimization problem (\ref{eq6}) is converted to the  problem (\ref{eq10}). To solve Eq. (\ref{eq10}), we use the following lemma and theorem.

\noindent\textbf{Lemma 1.
	\cite{re20}} Let $\bm{Y}=\bm{U} \bm{\Sigma} \bm{V}^{\top}$ be the SVD of $\bm{Y}\in{\mathbb{R}^{{n_1}\times{n_2}}}$, $r=\min(n_1,n_2)$, $\bm{\omega}=(\omega_1, \omega_2,\ldots,\omega_r)$ satisfies $0 \leq \omega_1 \leq \omega_2 \leq \cdots \leq \omega_r$. For any $\tau>0$, a global optimal solution of the following minimization objective
\begin{equation}
	\min_{\bm{X}} \tau\sum_{j=1}^{r}  \omega_{j} \sigma_{j}(\bm{X})+\frac{1}{2}\|\bm{X}-\bm{Y}\|_{F}^{2}
\end{equation}
is given by WSVT, i.e., 
\begin{equation}
	\bm{X}^*=\bm{U} \bm{S}_{\tau \bm{\omega}}(\bm{\Sigma}) \bm{V}^{\top},
\end{equation}
where $\bm{S}_{\tau\bm{\omega}}(\bm{\Sigma})=\big(\bm{\Sigma}-\tau \text{diag}(\bm{\omega})\big)_+$, and $(x)_+$ representes the positive part of $x$, i.e., $(x)_+=\max(x,0)$.

\noindent\textbf{Theorem 3.}
Let $\bm{\mathcal{Y}}=\bm{\mathcal{U}}*\bm{\mathcal{S}}*\bm{\mathcal{V}}^\top$ be the t-SVD of $\bm{\mathcal{Y}}\in{\mathbb{R}^{{n_1}\times{n_2}\times{n_3}}}$, $\bm{\mathcal{W}}\in{\mathbb{R}^{{n_1}\times{n_2}\times{n_3}}}$ is a f-diagonal tensor whose $i$th frontal slice is diag$(w_{1}^{i}, w_{2}^{i},\ldots, w_{r}^{i})$,  where $r=\min(n_1,n_2)$ and  $0 \leq \omega_1^i \leq \omega_2^i \leq \cdots \leq \omega_r^i$. For any $\tau>0$, a global
optimal solution of the following minimization objective
\begin{equation}
	\min_{\bm{\mathcal{X}}} \tau \frac{1}{n_3}\sum_{i=1}^{n_3} \sum_{j=1}^{r} \omega_{j}^{i} \sigma_{j}(\bar{\bm{X}}^{(i)})+\frac{1}{2}\|\bm{\mathcal{X}}-\bm{\mathcal{Y}}\|_{F}^{2}
	\label{eq11}
\end{equation}
is given by the tensor WSVT (t-WSVT)
\begin{equation}
	\bm{\mathcal{X}}^*=\bm{\mathcal{U}} * \bm{\mathcal{S}}_{\footnotesize{\tau \bm{\mathcal{W}}}} * \bm{\mathcal{V}}^\top,
\end{equation}
where $\bm{\mathcal{S}}_{\footnotesize{\tau \bm{\mathcal{W}}}}=\text{ifft}\big((\bar{\bm{\mathcal{S}}}-\tau
\bm{\mathcal{W}})_+, [\,], 3\big)$.

\noindent\textbf{Proof:}
In Fourier domain, the problem (\ref{eq11}) is equivalent to
\begin{align}
	&\min_{\bm{\mathcal{X}}} \tau \frac{1}{n_3}\sum_{i=1}^{n_3} \sum_{j=1}^{r} \omega_{j}^{i} \sigma_{j}(\bar{\bm{X}}^{(i)})+\frac{1}{2n_3}\left\|\bar{\bm{X}}-\bar{\bm{Y}}\right\|_{F}^{2},\\
	&=\min_{\bm{\mathcal{X}}}\frac{1}{n_3} \sum_{i=1}^{n_{3}} \,\Big(\tau\sum^{r}_{j=1}\omega_{j}^{i}\sigma_{j}(\bar{\bm{X}}^{(i)})+\frac{1}{2}\left\|\bar{\bm{X}}^{(i)}-\bar{\bm{Y}}^{(i)}\right\|_{F}^{2} \Big).
	\label{eq12}
\end{align}

In Eq. (\ref{eq12}), the variables $\bar{\bm{X}}^{(i)}$ are independent. Then, the above problem can be
divided into $n_3$ independent subproblems. By Lemma 1, we know  that the global optimal solution of the $i$th $(i=1,2,\ldots,n_{3})$ subproblem is the $i$th frontal slice of $\bar{\bm{\mathcal{X}}}^*$. Thus, $\bm{\mathcal{X}}^*$ is the solution of the problem (\ref{eq12}).

According to Eq. (\ref{eq9}) and Theorem 2, the solution of the problem (\ref{eq10}) is 
\begin{equation}
	\bm{\mathcal{L}}_{k+1}=\bm{\mathcal{U}} * \bm{\mathcal{S}}_{\footnotesize{\mu_k^{-1}\bm{\mathcal{W}}}} * \bm{\mathcal{V}}^\top,
	\label{eq13}
\end{equation}
where  $\bm{\mathcal{Q}}_k=\bm{\mathcal{U}}*\bm{\mathcal{S}}*\bm{\mathcal{V}}^\top$ is the t-SVD of $\bm{\mathcal{Q}}_k$, $\bm{\mathcal{W}}$ is a f-diagonal tensor which the $i$th frontal slice is diag$(w_{1,k}^{i}, w_{2,k}^{i},\ldots, w_{r,k}^{i})$. 
According to the weight  tensor $\bm{\mathcal{W}}$, Eq. (\ref{eq13}) utilizes small thresholds to shrink the large singular values and large thresholds to shrink the small singular values.

Similarly, holding  $\bm{\mathcal{L}}_{k+1}$, $\bm{\mathcal{P}}_k$, $\mu_k$ fixed, $\bm{\mathcal{E}}_{k+1}$ can be updated by
\begin{equation}\label{update E}
	\bm{\mathcal{E}}_{k+1}= \arg\min_{\bm{\mathcal{E}}}\, \lambda \|\bm{\mathcal{E}}\Vert_1+\frac{\mu_k}{2}\|\bm{\mathcal{E}}-\bm{\mathcal{H}}_k\Vert^2_F,\\
\end{equation}
where $\bm{\mathcal{H}}_k=\bm{\mathcal{X}}-\bm{\mathcal{L}}_{k+1}-\mu_k^{-1} \bm{\mathcal{P}}_k$. It has the following closed-form solution 
\begin{equation}
	\bm{\mathcal{E}}_{k+1}=\text{D}_{\lambda \cdot \mu_k^{-1}}\big(\bm{\mathcal{H}}_k\big),
	\label{eq14}
\end{equation}
where $\text{D}_{\tau}(x)$ is the soft thresholding operator \cite{re49} defined as
\begin{equation}
	\text{D}_{\tau}(x)= \begin{cases}0 & \text {if} \quad|x| \leq \tau, \\ \text{sign}(x)(|x|-\tau) & \text {if} \quad|x|>\tau.\end{cases}
\end{equation}

The whole optimization procedure for our N-TRPCA method is summarized in Algorithm 1.

\begin{table}[h]
	\label{Alg1}  
	\centering
	\begin{tabularx}{0.94\linewidth}{l}
		\toprule
		\textbf{Algorithm 1. N-TRPCA} \\
		\midrule
		\textbf{Input:} corrupted  tensor $\bm{\mathcal{X}}$.\\
		\textbf{Output:} recovered tensor $\hat{\bm{\mathcal{L}}}$.\\
		\textbf{Initialize:}  $\bm{\mathcal{L}}_0=\bm{\mathcal{E}}_0=\bm{\mathcal{P}}_0=0$, $\lambda$, $\mu_0$, $\mu_{\max}$,  $\rho$, $\epsilon$.\\
		\textbf{while} not converged \textbf{do}\\
		1. Update $\bm{\mathcal{L}}_{k+1}$ via (\ref{eq13});\\
		2. Update $\bm{\mathcal{E}}_{k+1}$ via (\ref{eq14});\\
		3. Update $\bm{\mathcal{P}}_{k+1}$ via $\bm{\mathcal{P}}_{k+1}=\bm{\mathcal{P}}_k+\mu_k\big(\bm{\mathcal{L}}_{k+1}+\bm{\mathcal{E}}_{k+1}$\\ \quad $-\bm{\mathcal{X}}\big)$;\\
		4. Update $\mu_{k+1}$ via $\mu_{k+1}=\min(\rho\mu_k,\mu_{\max})$;\\
		5. Check the convergence conditions\\
		\quad $\|\bm{\mathcal{L}}_{k+1}-\bm{\mathcal{L}}_k\Vert_{\infty}\leq\epsilon$, $\|\bm{\mathcal{E}}_{k+1}-\bm{\mathcal{E}}_k\Vert_{\infty}\leq\epsilon$, \\ \quad
		$\|\bm{\mathcal{L}}_{k+1}+\bm{\mathcal{E}}_{k+1}-\bm{\mathcal{X}}\Vert_{\infty}\leq\epsilon$;\\
		\textbf{end while} \\
		6. Let $\hat{\bm{\mathcal{L}}}=\bm{\mathcal{L}}_{k+1}$ be the recovered tensor.\\
		\bottomrule
	\end{tabularx}
\end{table}

\begin{figure*}[!t]
	\centering
	\includegraphics[width=0.78\linewidth]{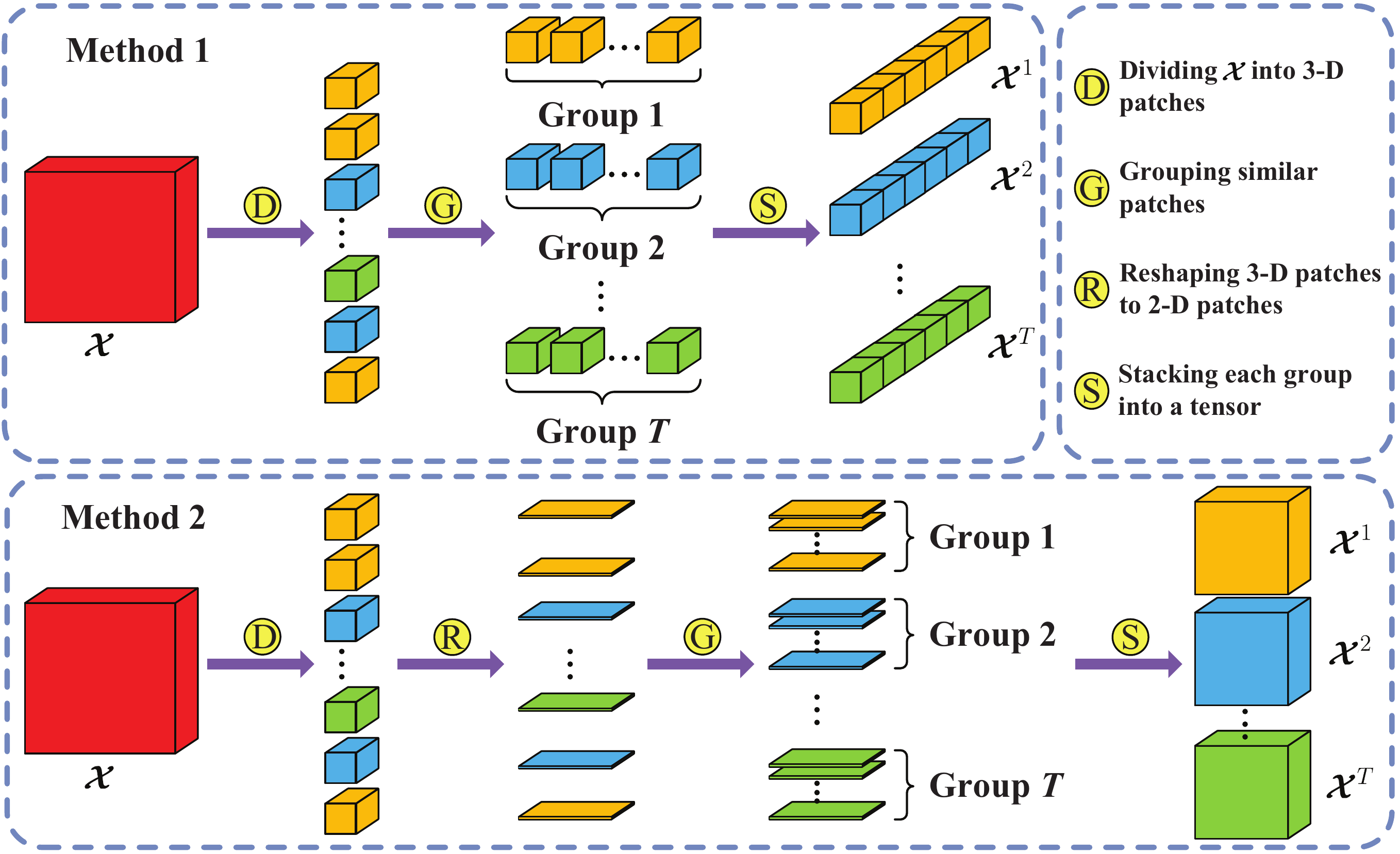}
	\caption{An illustration of two methods to construct group tensors.}
	\label{Fig_group}
\end{figure*}

\subsection{NN-TRPCA Model}\label{sec4.3}

As discussed in Section \ref{sec3.2}, nonlocal self-similarity implies that there is a lot of nonlocal structure redundancy in visual data. Therefore, for each patch of visual data tensor, we can search a group of similar tensor patches. Due to the high correlation between similar patches, the tensor formed by stacking these similar patches is low-rank. Therefore, we can apply our N-TRPCA to each formed tensor to obtain the final recovered visual data tensor. This inspires us to develop a nonlocal variant of N-TRPCA, i.e., NN-TRPCA. In the following, we will elaborate the proposed NN-TRPCA model. Its procedure mainly consists of three stages: tensor patch grouping, tensor patch low-rank recovery by N-TRPCA, and tensor patch aggregation.

\textbf{Tensor patch grouping}: Given a corrupted tensor $\bm{\mathcal{X}}\in{\mathbb{R}^{{n_1}\times{n_2}\times{n_3}}}$, we  divide $\bm{\mathcal{X}}$ into overlapping tensor patches with the spatial size $p\times p$. We then consider two methods to construct group tensors. See Fig. \ref{Fig_group} for an intuitive illustration of these two methods. Method 1 constructs each group tensor by stacking similar patches along 3-rd dimension. Specifically, for each 3-D patch, we search its $m-1$ similar patches based on the Euclidean metric. The group of each reference patch and its similar patches is indicated as $\Psi^t=\{{\bm{\mathcal{Y}}_i}\in{\mathbb{R}^{p \times p \times {n_3}}},i=1,2,\ldots,m\}, \, (t=1,2,\ldots,T)$. Here, $T$ is the number of groups. At last, the group  $\Psi^t$ is stacked into a third-order tensor $\bm{\mathcal{X}}^t \in \mathbb{R}^{p\times p \times m{n_3}}$. Differently, Method 2 unfolds tensor patches to matrix patches and stacks similar patches along the first dimension. Specifically, we first reshape all 3-D patches to 2-D patches with size $p^2 \times {n_3}$. For each 2-D patch, we find its $m-1$ similar patches. This similar group is denoted as $\Psi^t=\{{\bm {Y}_i}\in{\mathbb{R}^{p^2 \times {n_3}}},i=1,2,\ldots,m\}, \, (t=1,2,\ldots,T)$. The last step in Method 2 is to stack the group $\Psi^t$ into a third-order tensor $\bm{\mathcal{X}}^t \in \mathbb{R}^{m\times p^2 \times {n_3}}$. Note that, $\bm{\mathcal{X}}^t$ constructed by these two methods are both strongly low-rank because the similar patches in each group have strong correlations.

\textbf{Tensor patch low-rank recovery by N-TRPCA}: After nonlocal similar patches are grouped as $\bm{\mathcal{X}}^t, \, (t=1,2,\ldots,T)$, the low-rank tensor $\bm{\mathcal{L}}^t$ is estimated from $\bm{\mathcal{X}}^t$ by the N-TRPCA model, which is depicted to solve the following optimization problem
\begin{equation}
	\begin{split}
		&\min_{{\bm{\mathcal{L}}^t},{\bm{\mathcal{E}}^t}} \|\bm{\mathcal{L}}^t\Vert_{log}+ \lambda\|\bm{\mathcal{E}}^t\Vert_1,
		\\ s.t.\,&\bm{\mathcal{X}}^t=\bm{\mathcal{L}}^t+\bm{\mathcal{E}}^t,\, (t=1,2,\ldots,T).
	\end{split}
	\label{eq16}
\end{equation}

\textbf{Tensor patch aggregation}: At last, we reconstruct $\bm{\mathcal{L}}^t\, (t=1,2,\ldots,T)$ to its original position in tensor $\bm{\mathcal{X}}$ for obtaining the final recovered tensor $\hat{\bm{\mathcal{L}}}$. Note that, a pixel in overlapping regions of patches has multiple estimated values, we perform an averaging of them to obtain the final estimate.

Here, we discuss Method 1 and Method 2 of the first stage. Intuitively, if NN-TRPCA is applied to visual data restoration, adopting Method 1 to construct the group tensors will achieve better restoration performance than using Method 2. The reason is that the unfolding operators in Method 2 will destroy the structural information of tensor patches. To demonstrate our intuition, we conduct color image restoration experiments on 20 images randomly selected from the Berkely Segmentation Dataset \cite{re51}. The average quantitative results are tabulated in Table~\ref{tab_1}. From them, we can observe that whether Method 1 or Method 2 is adopted, NN-TRPCA outperforms N-TRPCA with respect to all the evaluation indices. This indicates that introducing the nonlocal prior is effective for visual data restoration. But using Method 2 obtains better restoration results than using Method 1. This observation is contrary to our intuition. One reason is that the size of tensor patches is small, so the loss of structure information caused by unfolding operators can be negligible. More importantly, these two methods stack similar patches along different dimensions. Method 1 uses FFT to capture the similarity between patches while Method 2 uses SVD. Thus, the other reason is that the data compression capability of SVD is better than that of FFT. In this work, for achieving better restoration performance, we choose Method 2 to form the group tensors.

\begin{table}[!t]
	\centering
	\caption{Quantitative performance comparison of different methods on 20 images with noise rate $30\%$. The evaluation indices are PSNR, SSIM and FSIM. }
	\begin{tabular}{m{3.9 cm}<{\centering}|m{1 cm}<{\centering} m{1 cm}<{\centering} m{1 cm}<{\centering}}
		\toprule
		Model&PSNR&SSIM&FSIM\\
		\midrule
		N-TRPCA&27.78&0.826&0.905\\
		%\hline
		NN-TRPCA (Method 1)&29.35&0.876&0.917\\
		%\hline
		NN-TRPCA (Method 2)&\textbf{32.61}&\textbf{0.935}&\textbf{0.955}\\
		\bottomrule
	\end{tabular}
	\label{tab_1}
\end{table}

The overall procedure of the NN-TRPCA model is described in Algorithm 2.

\begin{table}[h]
	\centering
	\begin{tabular}{l}
		\toprule
		\textbf{Algorithm 2. NN-TRPCA} \\
		\midrule
		\textbf{Input:} corrupted tensor $\bm{\mathcal{X}}$.\\
		\textbf{Output:} recovered tensor $\hat{\bm{\mathcal{L}}}$.\\
		1.  $\{\Psi^t\}_{t=1}^T$ $\leftarrow$ Divide the nonlocal similar patches of  \\ \quad tensor $\bm{\mathcal{X}}$ into $T$ groups;\\		
		2. $\bm{\mathcal{X}}^t$ $\leftarrow$ Stack the similar patches in group $\Psi^t$ as a \\ \quad tensor;\\
		3. \textbf{for} $t=1$ to $T$ \textbf{do}\\
		4. $\bm{\mathcal{L}}^t$ $\leftarrow$ Solve Eq. (\ref{eq16}) on $\bm{\mathcal{X}}^t$ via Algorithm 1;\\
		5. \textbf{end for}\\
		6. $\hat{\bm{\mathcal{L}}}$ $\leftarrow$ Output the recovered tensor by aggregating \\ \quad all $\bm{\mathcal{L}}^t$ $(t=1,2,\ldots,T)$ to the original position in $\bm{\mathcal{X}}$.\\
		\bottomrule
	\end{tabular}
	\label{Alg2}  
\end{table}

\subsection{Convergent Analysis of Algorithm 1}
\noindent\textbf{Lemma 2. \cite{re81}}\label{lewis}
Suppose $F: \mathbb{R}^{n_1\times n_2}\rightarrow \mathbb{R}$ is denoted as $F(\bm{X})=f \circ \sigma(\bm{X})$, $f$ is differentiable, and let 
$\bm{X}=\bm{U} \bm{\Sigma} \bm{V}^{\text{T}}$ be the SVD of $\bm{X}\in\mathbb{R}^{n_1\times n_2} $, $\bm{\Sigma}=\text{diag}(\sigma_1, \ldots, \sigma_n)$, $n=\min(n_1, n_2)$. The gradient of $F(\bm{X})$ at $\bm{X}$ is
\begin{equation}
	\label{deritheorem}
	\frac{\partial F(\bm{X})}{\partial \bm{X}}=\bm{U} \text{diag}(\theta) \bm{V}^{\text{T}},
\end{equation}
where $\theta=\frac{\partial f(y)}{\partial y}|_{y=\sigma (\bm{X})}$.

\noindent\textbf{Theorem 4.}\label{KKT}
The tensors $\bm{\mathcal{L}}_k$, $\bm{\mathcal{E}}_k$, and $\bm{\mathcal{P}}_k$ outputed by the proposed Algorithm 1 are bounded. And $\{\bm{\mathcal{L}}^\ast,\bm{\mathcal{E}}^\ast,\bm{\mathcal{P}}^\ast\}$ is a KKT stationary point of the objective \eqref{eq5}, satisfying the KKT conditions
$\bm{\mathcal{P}}^\ast\in\partial\|\bm{\mathcal{L}}\|_{log},\ \bm{\mathcal{L}}^\ast+\bm{\mathcal{E}}^\ast=\bm{\mathcal{X}},\ \bm{\mathcal{P}}^\ast\in\partial \|\bm{\mathcal{E}}\|_{1}.$

\noindent\textbf{Proof:}
$\bm{\mathcal{E}}_{k+1}$ satisfies the first-order necessary local optimality condition of \eqref{update E},
\begin{equation}
	\begin{split}
		\label{bound1}
		&0\in\partial_{\bm{\mathcal{E}}} L\left(\bm{\mathcal{L}}_{k+1},\bm{\mathcal{E}}_{k+1}, \bm{\mathcal{P}}_{k},\mu_{k}\right)\\
		=&\partial\left(\lambda \|\bm{\mathcal{E}}_{k+1}\|_{1}\right)+\bm{\mathcal{P}}_{k}+\mu_{k}\left(\bm{\mathcal{L}}_{k+1}-\bm{\mathcal{X}}+\bm{\mathcal{E}}_{k+1}\right)\\
		%=&U\Sigma_{k+1}V^T+\mu_{k}(J_{k+1}-Z_{k+1}+Z_{k+1}-Z_{k}+\frac{Y_2_{k}}{\mu_{k}})\\
		=&\partial \left(\lambda\|\bm{\mathcal{E}}_{k+1}\|_{1}\right)+\bm{\mathcal{P}}_{k+1}.
	\end{split}
\end{equation}
Due to  $\bm{\|\mathcal{E}}\|_{1}$ being non-smooth at $\bm{\mathcal{E}}_{ijk}=0$, we redefine sub-gradient $\left[\partial\|\bm{\mathcal{E}}\|_{1}\right]_{ijk}=0$ if $\bm{\mathcal{E}}_{ijk}=0$. Then $0\leq \| \partial\|\bm{\mathcal{E}}\|_{1} \|_F^2\leq n_1n_2n_3 $, hence $\partial (\lambda\|\bm{\mathcal{E}}_{k+1}\|_{1})$ is bounded. Thus it is ready to see that $\bm{\mathcal{P}}_{k}$ is bounded.

Considering the objective \eqref{eq5}, we have
\begin{align}
	&L\left(\bm{\mathcal{L}}_{k},\bm{\mathcal{E}}_{k},\bm{\mathcal{P}}_{k},\mu_{k}\right)
	\nonumber\\&=L\left(\bm{\mathcal{L}}_{k},\bm{\mathcal{E}}_{k},\bm{\mathcal{P}}_{k-1},\mu_{k-1}\right)+\frac{\mu_{k}-\mu_{k-1}}{2}\|\bm{\mathcal{L}}_{k}-\bm{\mathcal{X}}+\bm{\mathcal{E}}_{k}\|_F^2 \nonumber \\
	&+tr[(\bm{\mathcal{P}}_{k}-\bm{\mathcal{P}}_{k-1})(\bm{\mathcal{L}}_{k}-\bm{\mathcal{X}}+\bm{\mathcal{E}}_{k})],
	\nonumber\\&=L\left(\bm{\mathcal{L}}_{k},\bm{\mathcal{E}}_{k},\bm{\mathcal{P}}_{k-1},\mu_{k-1}\right)+\frac{\mu_{k}+\mu_{k-1}}{2(\mu_{k-1})^2}\|\bm{\mathcal{P}}_{k}-\bm{\mathcal{P}}_{k-1}\|_F^2.
\end{align}
So,
\begin{align}
	&L\left(\bm{\mathcal{L}}_{k+1},\bm{\mathcal{E}}_{k+1},\bm{\mathcal{P}}_{k},\mu_{k}\right)\nonumber\\
	&\leq L(\bm{\mathcal{L}}_{k+1},\bm{\mathcal{E}}_{k}, \bm{\mathcal{P}}_{k},\mu_{k})
	\leq L(\bm{\mathcal{L}}_{k}, \bm{\mathcal{E}}_{k}, \bm{\mathcal{P}}_{k},\mu_{k})\nonumber\\
	&\leq L\left(\bm{\mathcal{L}}_{k},\bm{\mathcal{E}}_{k},\bm{\mathcal{P}}_{k-1},\mu_{k-1}\right)+\frac{\mu_{k}+\mu_{k-1}}{2(\mu_{k-1})^2}\|\bm{\mathcal{P}}_{k}-\bm{\mathcal{P}}_{k-1}\|_F^2.
\end{align}
An easy induction gives
\begin{align}&L(\bm{\mathcal{L}}_{k+1}, \bm{\mathcal{E}}_{k+1}, \bm{\mathcal{P}}_{k},\mu_{k}) \nonumber
	\\&\leq L\left(\bm{\mathcal{L}}_1, \bm{\mathcal{E}}_{1},\bm{\mathcal{P}}_{0},\mu_{0}\right)+\sum_{i=1}^{k}\frac{\mu_i+\mu_{i-1}}{2(\mu_{i-1})^2}\|\bm{\mathcal{P}}_i-\bm{\mathcal{P}}_{i-1}\|_F^2.
\end{align}

Since $\frac{\mu_i+\mu_{i-1}}{2(\mu_{i-1})^2}\|\bm{\mathcal{P}}_i-\bm{\mathcal{P}}_{i-1}\|_F^2$ is bounded,
it is not hard to know that $L\left(\bm{\mathcal{L}}_{k+1}, \bm{\mathcal{E}}_{k+1},\bm{\mathcal{P}}_{k},\mu_{k}\right)$ is upper bounded. And it is straightforward to show that
\begin{align}
	&L\left(\bm{\mathcal{L}}_{k+1}, \bm{\mathcal{E}}_{k+1},\bm{\mathcal{P}}_{k},\mu_{k}\right)+\frac{1}{2\mu_{k}}\|\bm{\mathcal{P}}_{k}\|_F^2
	\nonumber\\&=\|\bm{\mathcal{L}}_{k+1}\|_{log}+\!\lambda \|\bm{\mathcal{E}}_{k+1}\|_{1}+\!\frac{\mu_{k}}{2}\|\bm{\mathcal{L}}_{k+1}-\!\bm{\mathcal{X}}+\!\bm{\mathcal{E}}_{k+1}+\!\frac{\bm{\mathcal{P}_{k}}}{\mu_{k}}\|_F^2.
\end{align}
In the above equation, each term on the right-hand side is bounded, Therefore, $\bm{\mathcal{E}}_{k+1}$ is bounded. And $\bm{\mathcal{L}}_{k+1}$ is also bounded by the last term on the right-hand. 

According to Bolzano-Weierstrass theorem, we know that each infinite bounded sequence in $\mathbb {R}^{n}$ has a convergent subsequence. There must be at least one accumulation point of the sequence $\bm{\{\mathcal{L}}_k,\bm{\mathcal{E}}_k,\bm{\mathcal{P}}_k\}^{+\infty}_{k=1}$. We denote one of the points as $\{\bm{\mathcal{L}}^\ast,\bm{\mathcal{E}}^\ast,\bm{\mathcal{P}}^\ast\}$, and we assume $\{\bm{\mathcal{L}}_k,\bm{\mathcal{E}}_k,\bm{\mathcal{P}}_k\}^{+\infty}_{k=1}$ converge to $\{\bm{\mathcal{L}}^\ast,\bm{\mathcal{E}}^\ast,\bm{\mathcal{P}}^\ast\}$ without loss of generality.

\begin{table*}[!t]
	\begin{center}
		\caption{Quantitative Performance Comparison of different methods on 100 Test Images with Different Noise Rates.}
		\label{tab_result1}
		\begin{tabular}{m{2.8 cm}<{\centering} m{0.85 cm}<{\centering} m{0.85 cm}<{\centering} m{0.85 cm}<{\centering} m{0.3 cm}<{\centering} m{0.85 cm}<{\centering} m{0.85 cm}<{\centering} m{0.85 cm}<{\centering} m{0.3 cm}<{\centering} m{0.85 cm}<{\centering} m{0.85 cm}<{\centering} m{0.85 cm}<{\centering} m{0.1 cm}<{\centering}}
			\toprule
			\multirow{2}{*}{Method}& \multicolumn{3}{c}{$10\%$}&
			& \multicolumn{3}{c}{$20\%$} & & \multicolumn{3}{c}{$30\%$}&
			\\
			\cline{2-4} \cline{6-8} \cline{10-12} \specialrule{0em}{0pt}{4pt}
			&PSNR&SSIM&FSIM& &PSNR&SSIM&FSIM&  &PSNR&SSIM&FSIM&\\
			\midrule[0.4pt]
			SNN-TRPCA&26.09&0.812&0.869& &25.14&0.778&0.847& &23.11&0.735&0.822&\\
			KBR-TRPCA&29.50&0.919&0.958& &27.91&0.875&0.933& &25.97&0.793&0.892&\\
			TNN-TRPCA&29.35&0.935&0.945& &27.68&0.886&0.925& &25.73&0.794&0.878&\\
			TPSCPSF&30.56&0.898&0.940& &30.17&0.892&0.937& &30.09&0.884&0.933&\\
			N-TRPCA&34.31&0.957&0.977& &30.26&0.896&0.944& &27.20&0.809&0.898&\\
			NN-TRPCA&\textbf{38.91}&\textbf{0.980}&\textbf{0.988}& &\textbf{35.08}&\textbf{0.960}&\textbf{0.975}& &\textbf{31.79}&\textbf{0.926}&\textbf{0.952}&\\
			\bottomrule[0.4pt]
		\end{tabular}
	\end{center}
\end{table*}

Since $$(\bm{\mathcal{P}}_{k+1}-\bm{\mathcal{P}}_k)/\mu_k=\bm{\mathcal{L}}_{k+1}+\bm{\mathcal{E}}_{k+1}-\bm{\mathcal{X}},$$ we can deduce that $$\lim_{k\to\infty}(\bm{\mathcal{L}}_{k+1}+\bm{\mathcal{E}}_{k+1}-\bm{\mathcal{X}})=\lim_{k\to\infty}(\bm{\mathcal{P}}_{k+1}-\bm{\mathcal{P}}_k)/\mu_k=0.$$Then $\bm{\mathcal{L}}^\ast+\bm{\mathcal{E}}^\ast=\bm{\mathcal{X}}$ is achieved.

$\bm{\mathcal{L}}_{k+1}$ also satisfies the first-order necessary local optimality condition of (\ref{eq6})
$$0\in\partial\|\bm{\mathcal{L}}_{k+1}\|_{log}+\bm{\mathcal{P}}_k+\mu_k\left(\bm{\mathcal{L}}_{k+1}+\bm{\mathcal{E}}_{k}-\bm{\mathcal{X}}\right).$$ From Lemma 2 and Definition 6, for $i=1,...,n_3$, we have $$\nabla\|\bar{\bm{\mathcal{L}}}^{(i)}\|_{log}=\bar{\bm{U}}^{(i)}\mathrm{diag}(\frac{\theta}{\theta \sigma_j(\bar{\bm{\mathcal{L}}}^{(i)})+1})\bar{\bm{V}}^{(i)T},$$ and  $$\frac{\theta}{\theta \sigma_j(\bar{\bm{\mathcal{L}}}^{(i)})+1}< \theta\Longrightarrow \nabla\|\bar{\bm{\mathcal{L}}}^{(i)}\|_{log}$$ is bounded.
Thus, 
$
\frac{\partial\|\bm{\mathcal{L}}\|_{log}}{\partial\bar{\bm{\mathcal{L}}}}
$   
is bounded.
In fact, $\bar{\bm{\mathcal{L}}}$ is equivalent to Tucker product (see~\cite{re82}), $\bar{\bm{\mathcal{L}}}=\bm{\mathcal{L}}\times_3 \widetilde{\bm{\mathbf{F}}}_{n_3}.$ From it and using the chain rule, we can obtain that $$\nabla\|\bm{\mathcal{L}}\|_{log}=\frac{\partial \|\bm{\mathcal{L}}\|_{log}}{\partial\bar{\bm{\mathcal{L}}}}\times_3 \widetilde{\bm{\mathbf{F}}}_{n_3}^*$$ is bounded. And it is easy to know 
$$0\in\partial\|\bm{\mathcal{L}}_{k+1}\|_{log}+\bm{\mathcal{P}}_{k+1}-\mu_k\left(\bm{\mathcal{E}}_{k+1}-\bm{\mathcal{E}}_{k}\right).$$
Note that $\mu_k<\mu_{max}=10^{10}$ in Algorithm~1, then clearly $\lim_{k\to\infty}\mu_k\left(\bm{\mathcal{E}}_{k+1}-\bm{\mathcal{E}}_{k}\right)=0\Longrightarrow-\bm{\mathcal{P}}^\ast\in \partial\|\bm{\mathcal{L}}^\ast\|_{log}.$

Similarly, due to the fact that $\bm{\mathcal{E}}_{k+1}$ is the minimum of the subproblem $L\left(\bm{\mathcal{L}}_k,\bm{\mathcal{E}},\bm{\mathcal{P}}_k,\mu_k\right)$, we have
$$0\in\partial\|\bm{\mathcal{E}}_{k+1}\|_{1}+\bm{\mathcal{P}}_k+\mu_k\left(\bm{\mathcal{L}}_k+\bm{\mathcal{E}}_{k+1}-\bm{\mathcal{X}}\right).$$ So $\lim_{k\to\infty}\bm{\mathcal{P}}_{k+1}=\bm{\mathcal{P}}^\ast\in\partial\|\bm{\mathcal{E}}^\ast\|_{1}.$ Thus $\left(\bm{\mathcal{L}}^\ast,\bm{\mathcal{E}}^\ast,\bm{\mathcal{P}}^\ast\right)$ satisfies the Karush-Kuhn-Tuker (KKT) conditions of the Lagrange function $L(\bm{\mathcal{L}},\bm{\mathcal{E}},\bm{\mathcal{P}},\mu)$. We are now in a position to complete the proof.

\section{Experimental Results}\label{sec5}

We evaluate the performance of NN-TRPCA in the task of color image and gray video recovery. The color images and gray videos can be considered as third-order tensors, and the recovery task is to estimate the clean visual tensors from their corrupted versions.

\subsection{Experimental Setup}\label{sec5.1}

For color image restoration, we randomly select 100 color images with size $321\times481$ from the popular Berkely Segmentation Dataset \cite{re51} as test images. These images includes different natural scenes and objects, e.g., animals, plants, people, scenery, and buildings. For each color image, we vary the noise rate from 10 to 30 percent. The pixels of each rate are randomly set to random values in $[0,255]$, and the positions of corrupted pixels are unknown. All the three channels of a color image are corrupted in the same positions. This setting is more challenging than the noise of three channels in different positions. Besides, three video sequences from Scene Background Initialization Dataset \cite{re27} are choosen for video restoration, including `Hall $\&$ Monitor', `Candela$\_$m1.10', and  `CAVIAR1'. The frame sizes of these three videos are $ 352\times 240$, $ 352\times 288$,  and $ 384\times 256$, respectively. Owing to the computational limitation, we only use the first 30 frames of each video and resize each frame to a quarter of its original size. Similar to color images, for each gray video, the 30 percent of pixels is randomly set to random values in $[0,255]$, and the positions of corrupted pixels are also unknown.

\begin{figure*}[!t]
	\centering
	\subfloat{
		\begin{minipage}[ht]{0.11\linewidth}
			\centering
			\includegraphics[width=2.15 cm]{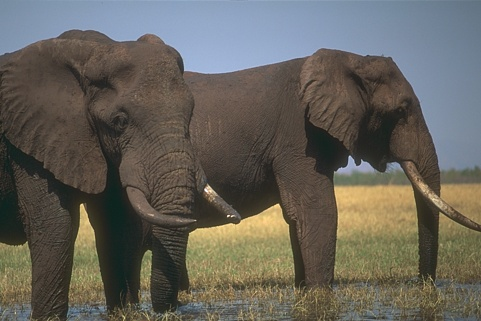}\\
			\vspace{0.07cm}
			\includegraphics[width=2.15 cm]{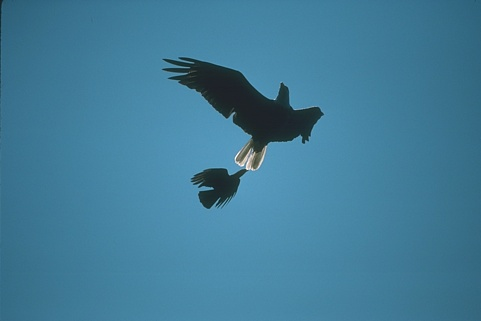}\\
			\vspace{0.07cm}
			\includegraphics[width=2.15 cm]{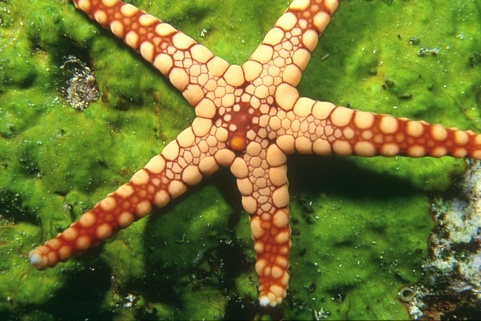}\\
			\vspace{0.07cm}
			\includegraphics[width=2.15 cm]{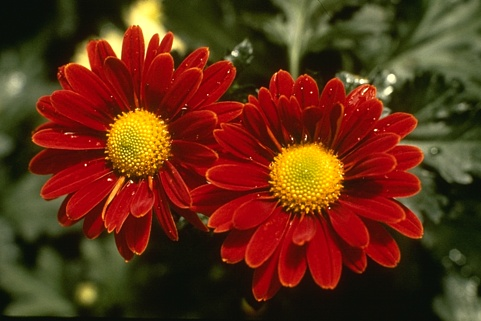}\\
			\vspace{0.07cm}
			\includegraphics[width=2.15 cm]{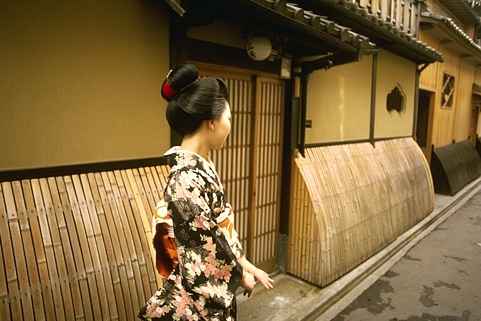}\\
			\vspace{0.07cm}
			\includegraphics[width=2.15 cm]{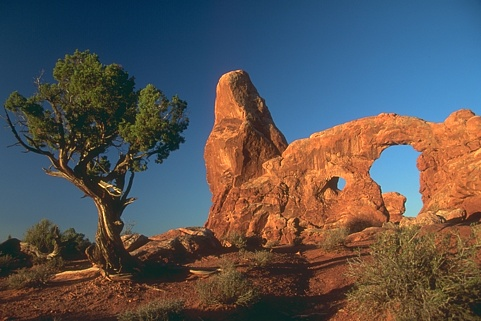}\\
			\vspace{0.07cm}
			\includegraphics[width=2.15 cm]{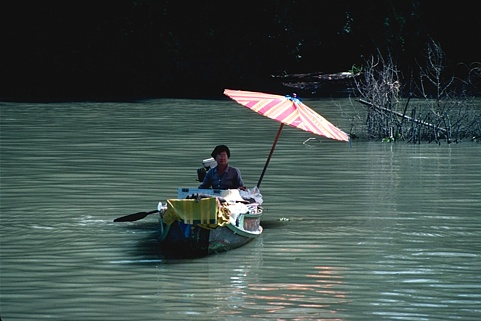}\\
			\vspace{0.07cm}
			\includegraphics[width=2.15 cm]{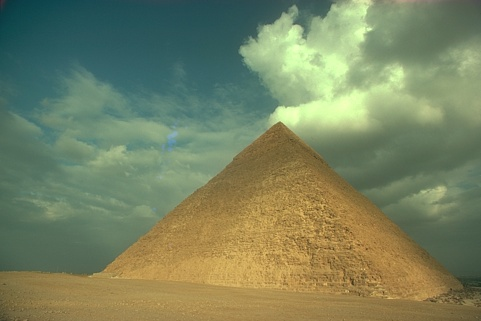}\\
			\centerline{\small Original}
		\end{minipage}
	} 	  
	\subfloat{
		\begin{minipage}[ht]{0.11\linewidth}
			\centering
			\includegraphics[width=2.15 cm]{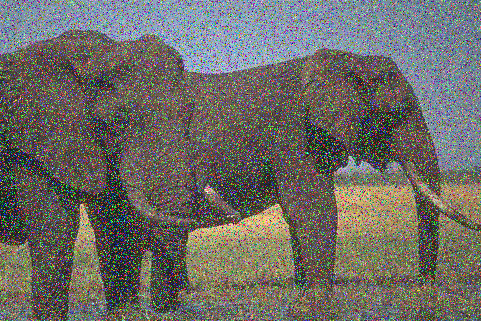}\\
			\vspace{0.07cm}
			\includegraphics[width=2.15 cm]{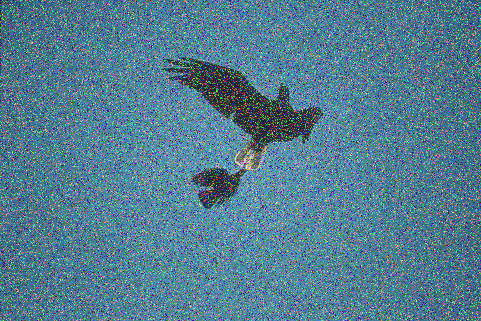}\\
			\vspace{0.07cm}
			\includegraphics[width=2.15 cm]{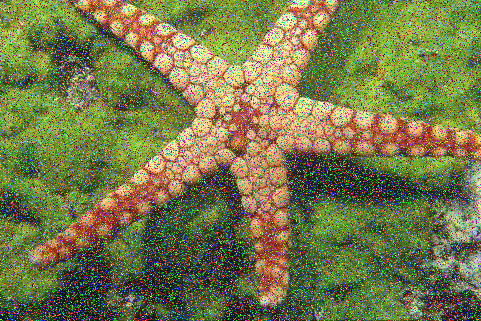}\\
			\vspace{0.07cm}
			\includegraphics[width=2.15 cm]{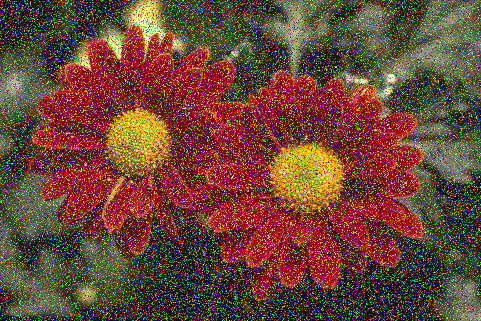}\\
			\vspace{0.07cm}
			\includegraphics[width=2.15 cm]{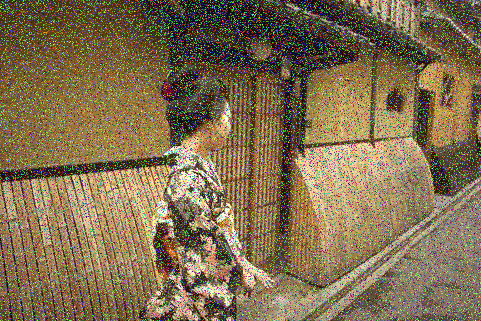}\\
			\vspace{0.07cm}
			\includegraphics[width=2.15 cm]{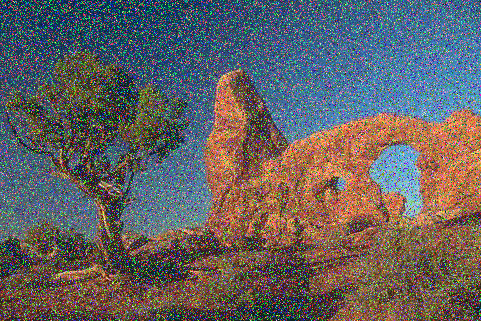}\\
			\vspace{0.07cm}
			\includegraphics[width=2.15 cm]{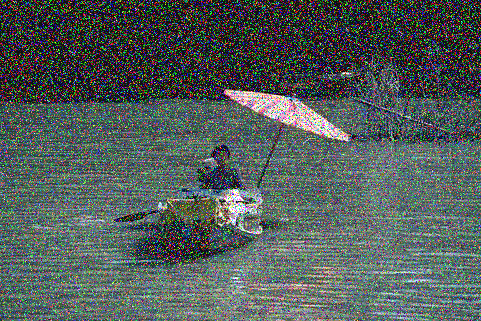}\\
			\vspace{0.07cm}
			\includegraphics[width=2.15 cm]{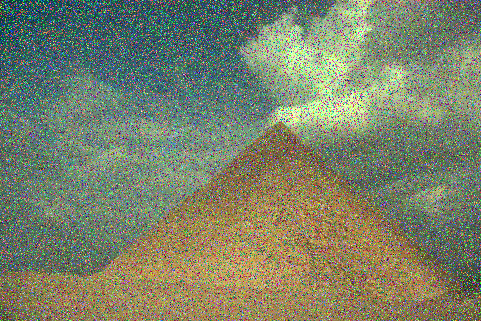}\\
			\centerline{\small Corrupted}
		\end{minipage}
	}
	\subfloat{
		\begin{minipage}[ht]{0.11\linewidth}
			\centering
			\includegraphics[width=2.15 cm]{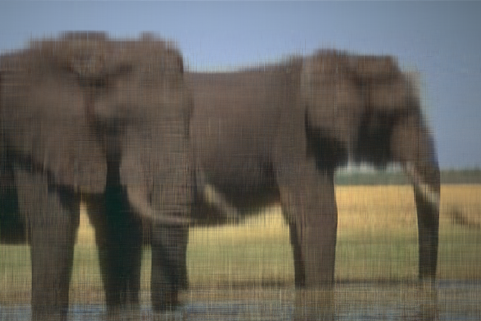}\\
			\vspace{0.07cm}
			\includegraphics[width=2.15 cm]{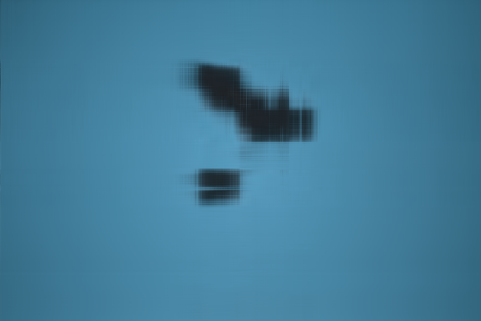}\\
			\vspace{0.07cm}
			\includegraphics[width=2.15 cm]{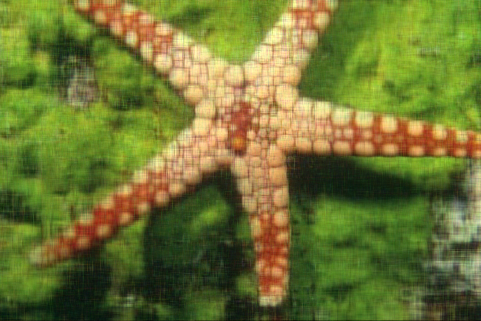}\\
			\vspace{0.07cm}
			\includegraphics[width=2.15 cm]{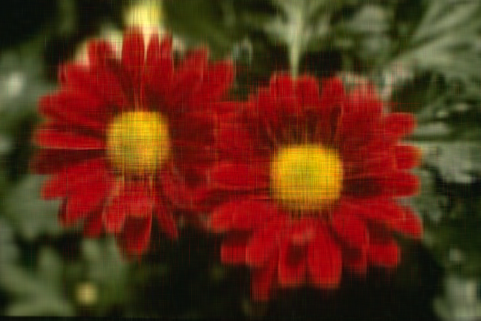}\\
			\vspace{0.07cm}
			\includegraphics[width=2.15 cm]{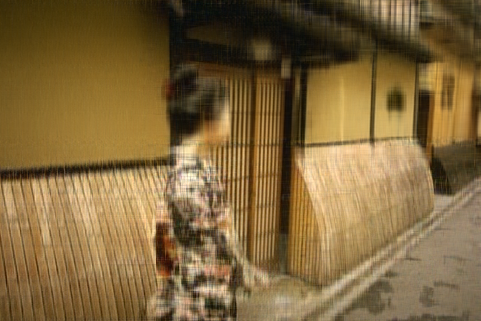}\\
			\vspace{0.07cm}
			\includegraphics[width=2.15 cm]{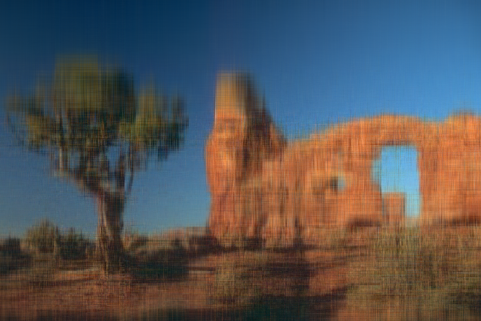}\\
			\vspace{0.07cm}
			\includegraphics[width=2.15 cm]{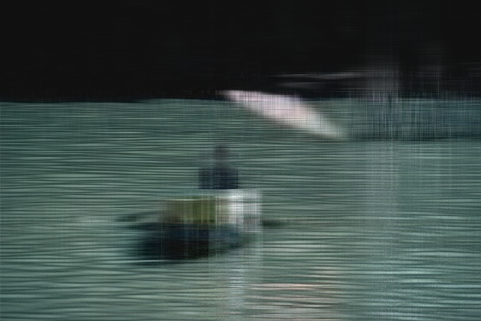}\\
			\vspace{0.07cm}
			\includegraphics[width=2.15 cm]{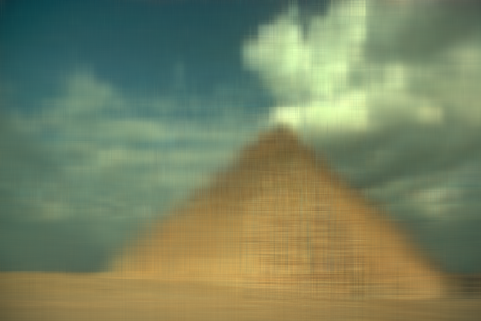}\\
			\centerline{\small SNN-TRPCA}
		\end{minipage}
	} 
	\subfloat{
		\begin{minipage}[ht]{0.11\linewidth}
			\centering
			\includegraphics[width=2.15 cm]{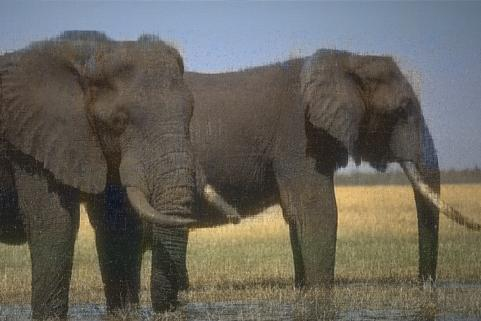}\\
			\vspace{0.07cm}
			\includegraphics[width=2.15 cm]{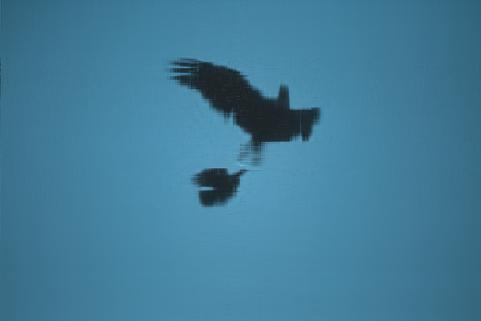}\\
			\vspace{0.07cm}
			\includegraphics[width=2.15 cm]{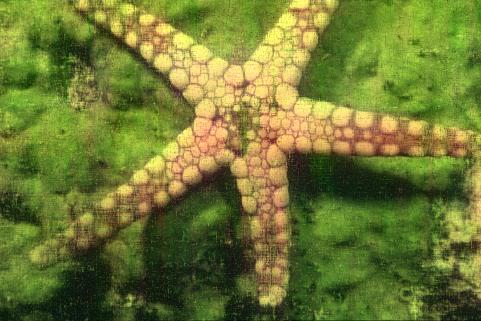}\\
			\vspace{0.07cm}
			\includegraphics[width=2.15 cm]{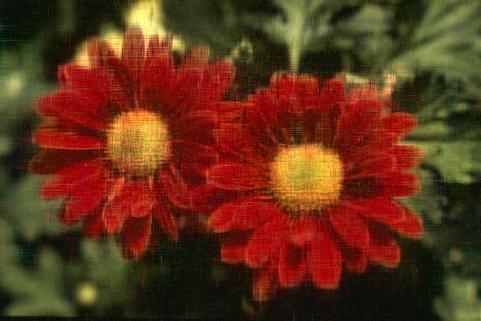}\\
			\vspace{0.07cm}
			\includegraphics[width=2.15 cm]{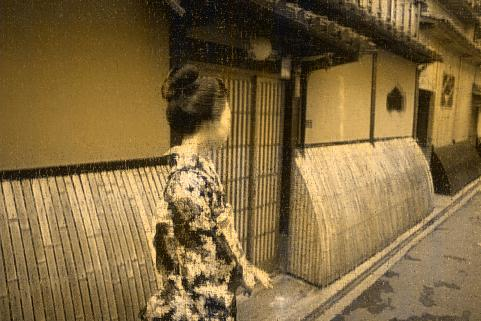}\\
			\vspace{0.07cm}
			\includegraphics[width=2.15 cm]{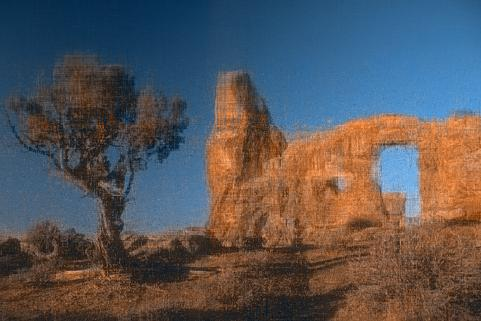}\\
			\vspace{0.07cm}
			\includegraphics[width=2.15 cm]{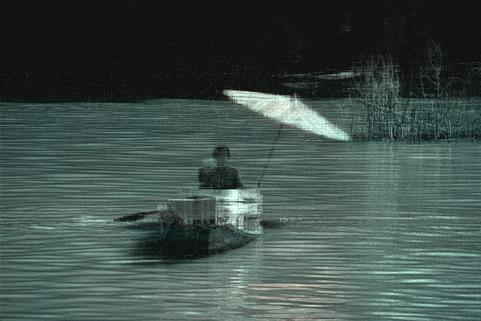}\\
			\vspace{0.07cm}
			\includegraphics[width=2.15 cm]{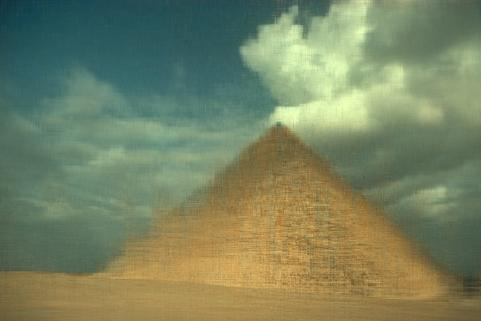}\\
			\centerline{\small KBR-TRPCA}
		\end{minipage}
	} 
	\subfloat{
		\begin{minipage}[ht]{0.11\linewidth}
			\centering
			\includegraphics[width=2.15 cm]{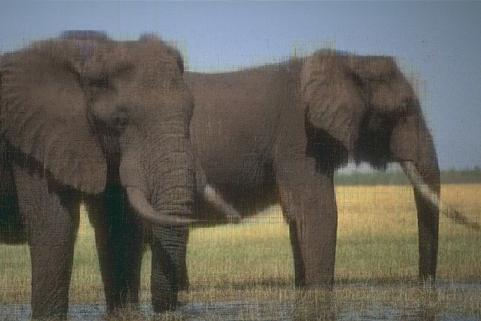}\\
			\vspace{0.07cm}
			\includegraphics[width=2.15 cm]{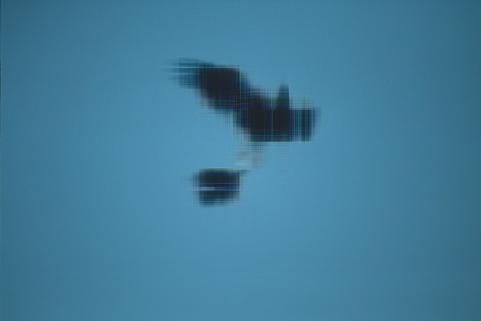}\\
			\vspace{0.07cm}
			\includegraphics[width=2.15 cm]{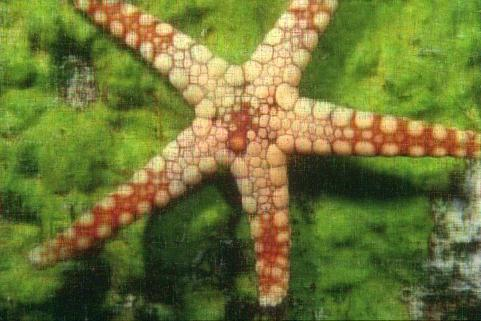}\\
			\vspace{0.07cm}
			\includegraphics[width=2.15 cm]{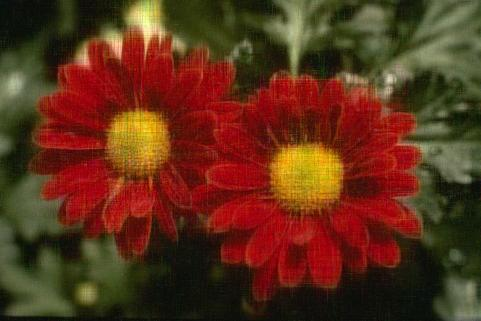}\\
			\vspace{0.07cm}
			\includegraphics[width=2.15 cm]{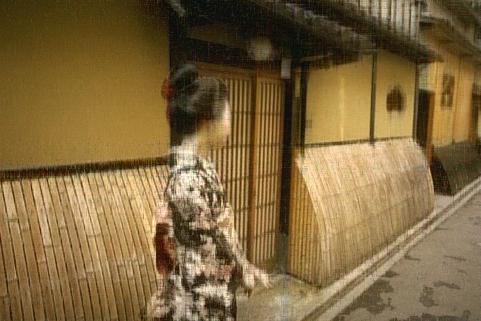}\\
			\vspace{0.07cm}
			\includegraphics[width=2.15 cm]{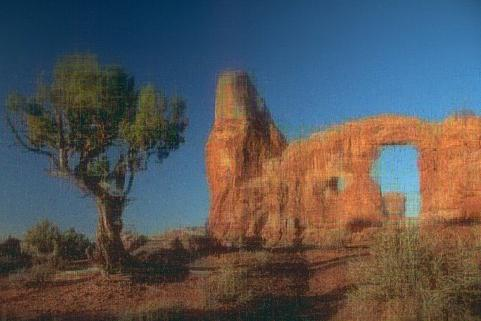}\\
			\vspace{0.07cm}
			\includegraphics[width=2.15 cm]{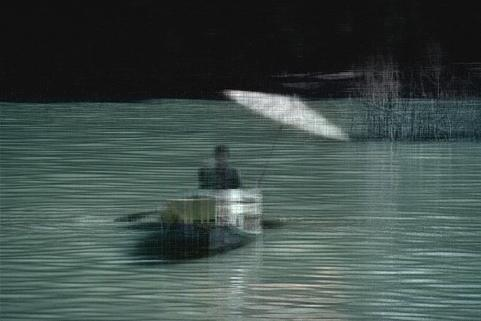}\\
			\vspace{0.07cm}
			\includegraphics[width=2.15 cm]{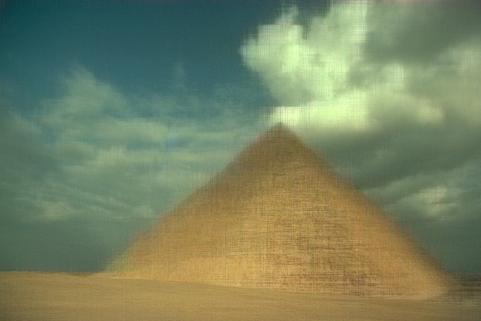}\\
			\centerline{\small TNN-TRPCA}
		\end{minipage}
	}
	\subfloat{
		\begin{minipage}[ht]{0.11\linewidth}
			\centering
			\includegraphics[width=2.15 cm]{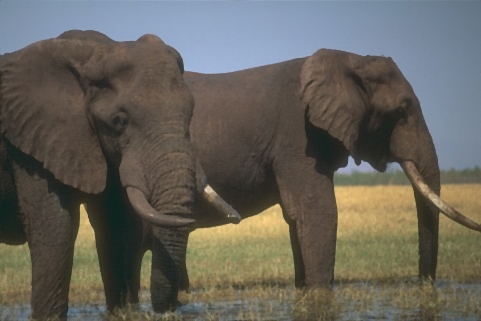}\\
			\vspace{0.07cm}
			\includegraphics[width=2.15 cm]{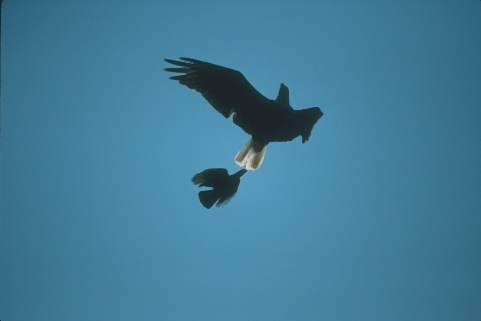}\\
			\vspace{0.07cm}
			\includegraphics[width=2.15 cm]{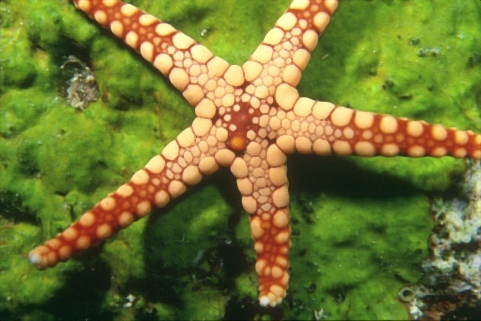}\\
			\vspace{0.07cm}
			\includegraphics[width=2.15 cm]{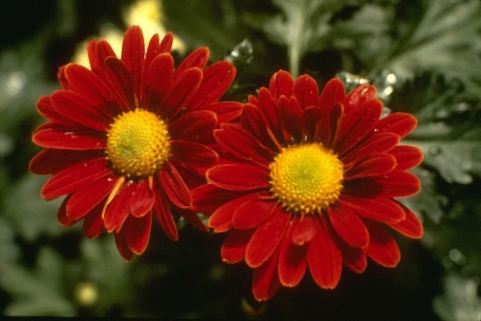}\\
			\vspace{0.07cm}
			\includegraphics[width=2.15 cm]{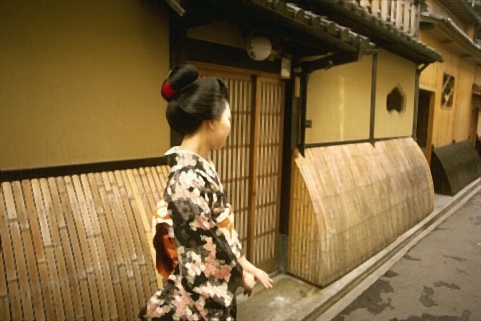}\\
			\vspace{0.07cm}
			\includegraphics[width=2.15 cm]{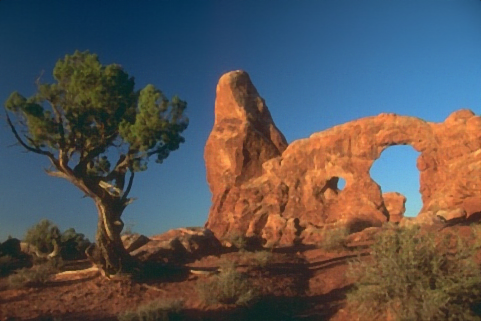}\\
			\vspace{0.07cm}
			\includegraphics[width=2.15 cm]{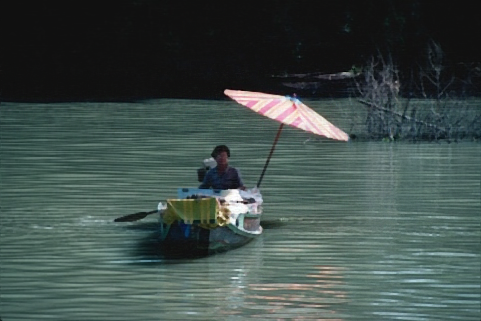}\\
			\vspace{0.07cm}
			\includegraphics[width=2.15 cm]{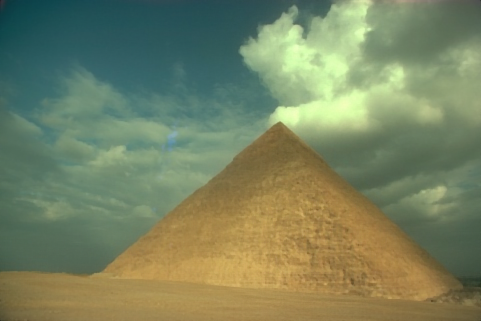}\\
			\centerline{\small TPSCPSF}
		\end{minipage}
	} 
	\subfloat{
		\begin{minipage}[ht]{0.11\linewidth}
			\centering
			\includegraphics[width=2.15 cm]{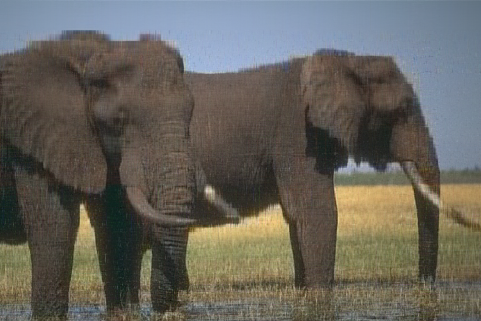}\\
			\vspace{0.07cm}
			\includegraphics[width=2.15 cm]{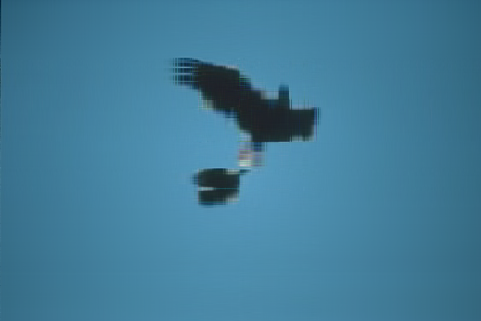}\\
			\vspace{0.07cm}
			\includegraphics[width=2.15 cm]{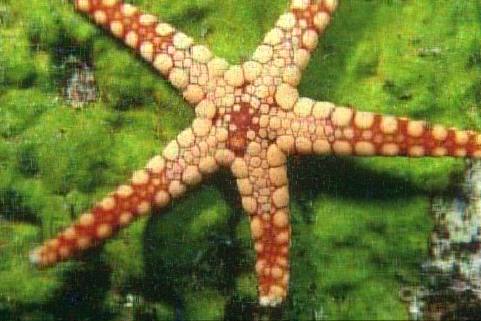}\\
			\vspace{0.07cm}
			\includegraphics[width=2.15 cm]{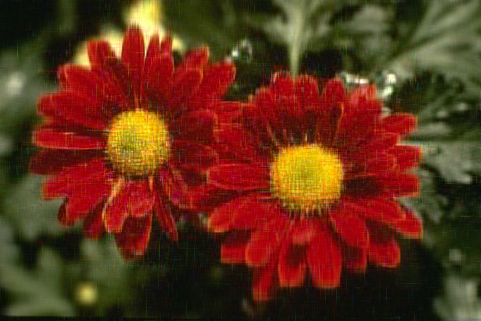}\\
			\vspace{0.07cm}
			\includegraphics[width=2.15 cm]{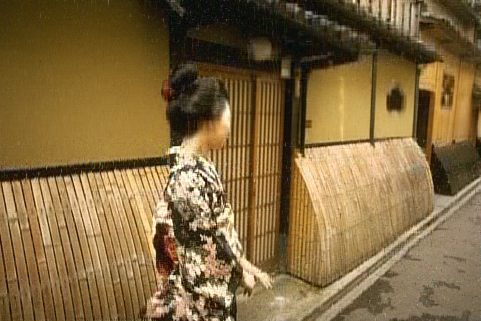}\\
			\vspace{0.07cm}
			\includegraphics[width=2.15 cm]{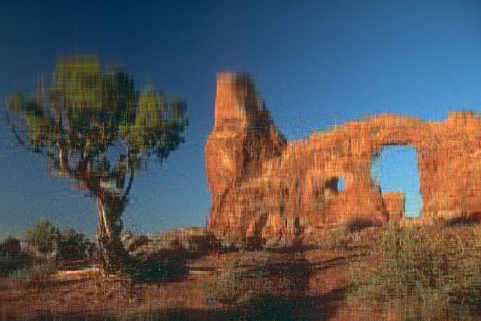}\\
			\vspace{0.07cm}
			\includegraphics[width=2.15 cm]{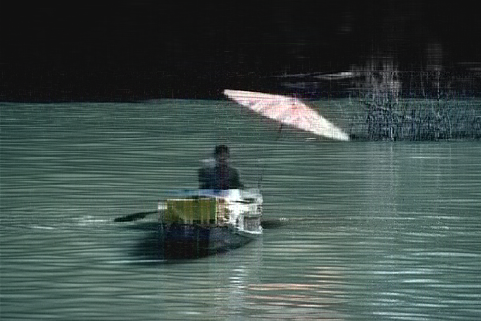}\\
			\vspace{0.07cm}
			\includegraphics[width=2.15 cm]{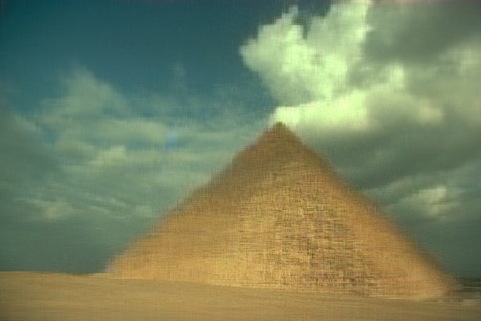}\\
			\centerline{\small N-TRPCA}
		\end{minipage}
	}
	\subfloat{
		\begin{minipage}[ht]{0.11\linewidth}
			\centering
			\includegraphics[width=2.15 cm]{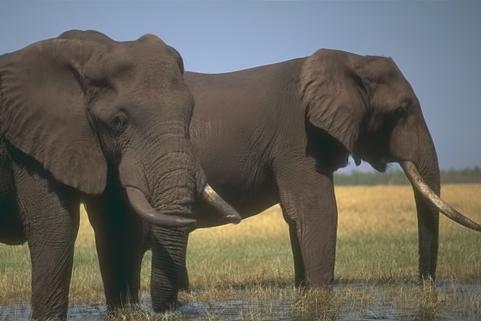}\\
			\vspace{0.07cm}
			\includegraphics[width=2.15 cm]{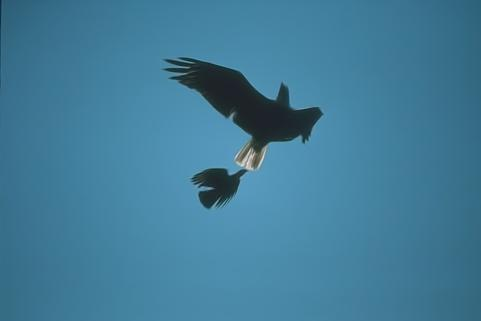}\\
			\vspace{0.07cm}
			\includegraphics[width=2.15 cm]{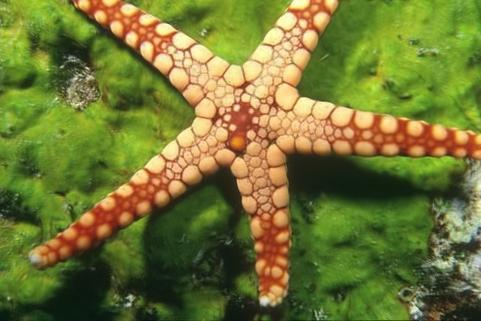}\\
			\vspace{0.07cm}
			\includegraphics[width=2.15 cm]{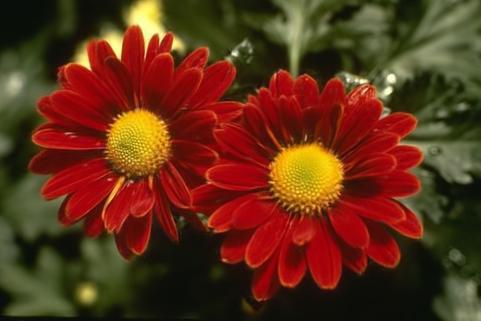}\\
			\vspace{0.07cm}
			\includegraphics[width=2.15 cm]{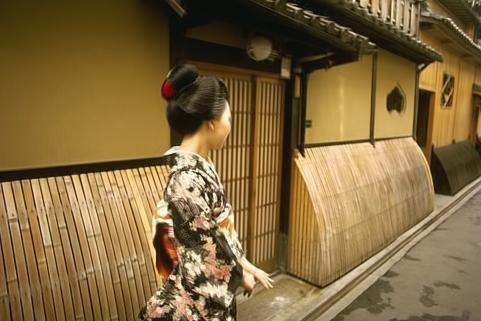}\\
			\vspace{0.07cm}
			\includegraphics[width=2.15 cm]{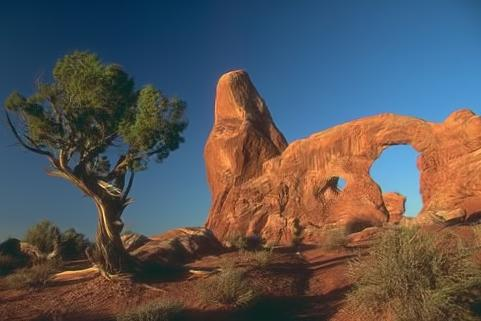}\\
			\vspace{0.07cm}
			\includegraphics[width=2.15 cm]{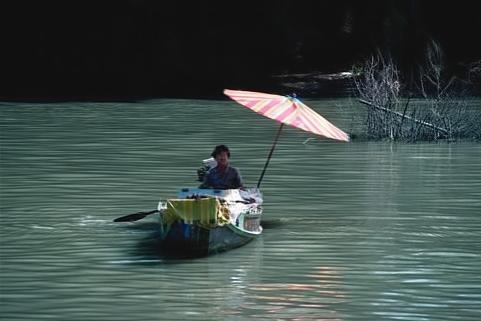}\\
			\vspace{0.07cm}
			\includegraphics[width=2.15 cm]{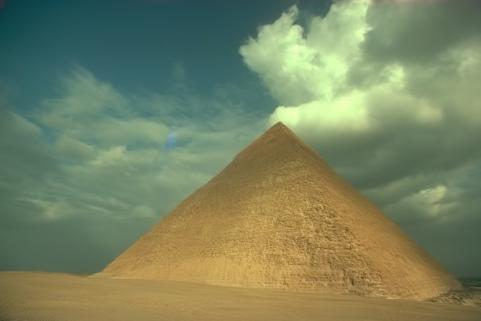}\\
			\centerline{\small NN-TRPCA}
		\end{minipage}
	} 
	\centering
	\caption{Restoration results on eight example images. From top to bottom: `Elephant', `Bird', `Starfish', `Flower', `Girl', `Stone', `Boat', and `Pyramid'. }
	\label{Fig_result1}
\end{figure*}

Our N-TRPCA and NN-TRPCA are compared with four state-of-the-art TRPCA methods, including SNN-TRPCA \cite{re11}, KBR-TRPCA \cite{re17}, TNN-TRPCA \cite{re5}, and TPCPSF \cite{re83}. These comparison methods adopt different tensor rank as low-rank constraints. SNN-TRPCA is based on the Tucker rank. KBR-TRPCA is a Kronecker-basis-representation based method that combines the Tucker rank and CP rank. TNN-TRPCA and TPCPSF are based on the tensor average rank. The performance of different methods are evaluated by three quantitative picture quality indices (PQI), including PSNR \cite{re5}, SSIM \cite{re24}, FSIM \cite{re25}. PSNR and SSIM are two commonly-used PQIs in the task of image restoration. The former measures the similarity between the ground-truth and the restored image based on MSE, and the latter is to measure the structural consistency. Unlike SSIM, FSIM is more consistent with human eye perception by utilizing both phase congruency and image gradient magnitude. The higher the values of these three indices, the better the restoration results.

\begin{table}%[!t]
	\centering
	\caption{Quantitative Performance Comparison of Different Methods on Eight Example Images with Noise Rate $30\%$.}
	\label{tab_result2}
	\begin{tabular}{m{1.2 cm}<{\centering} m{2.3 cm}<{\centering} m{0.7 cm}<{\centering} m{0.7 cm}<{\centering}  m{0.85 cm}<{\centering}}
		\toprule
		Image&Method&PSNR&SSIM&FSIM\\
		\midrule[0.4pt]
		\multirow{6}*{Elephant}&SNN-TRPCA&26.59&0.712&0.823\\
		&KBR-TRPCA&29.01&0.793&0.904\\
		&TNN-TRPCA&28.15&0.802&0.892\\
		&TPSCPSF&31.83&0.860&0.928\\
		&N-TRPCA&29.22&0.819&0.914\\
		&NN-TRPCA&\textbf{33.40}&\textbf{0.920}&\textbf{0.952}\\
		\midrule[0.4pt]
		\multirow{6}*{Bird}&SNN-TRPCA&29.10&0.949&0.892\\		
		&KBR-TRPCA&32.15&0.962&0.919\\
		&TNN-TRPCA&29.86&0.954&0.914\\
		&TPSCPSF&40.83&0.989&0.982\\
		&N-TRPCA&32.24&0.962&0.928\\
		&NN-TRPCA&\textbf{41.62}&\textbf{0.988}&\textbf{0.931}\\
		\midrule[0.4pt]
		\multirow{5}*{Starfish}&SNN-TRPCA&22.17&0.669&0.845\\
		&KBR-TRPCA&21.99&0.674&0.865\\
		&TNN-TRPCA&23.59&0.686&0.868\\
		&TPSCPSF&30.70&0.903&0.958\\
		&N-TRPCA&25.68&0.735&0.889\\
		&NN-TRPCA&\textbf{30.59}&\textbf{0.906}&\textbf{0.955}\\
		\midrule[0.4pt]
		\multirow{6}*{Flower}&SNN-TRPCA&24.34&0.721&0.864\\		
		&KBR-TRPCA&23.73&0.657&0.872\\
		&TNN-TRPCA&24.44&0.689&0.887\\
		&TPSCPSF&32.08&0.930&0.972\\
		&N-TRPCA&26.31&0.734&0.900\\
		&NN-TRPCA&\textbf{31.63}&\textbf{0.914}&\textbf{0.965}\\
		\midrule[0.4pt]
		\multirow{6}*{Girl}&SNN-TRPCA&21.71&0.756&0.810\\		
		&KBR-TRPCA&23.54&0.768&0.865\\
		&TNN-TRPCA&23.61&0.785&0.861\\
		&TPSCPSF&26.63&0.891&0.935\\
		&N-TRPCA&25.83&0.846&0.896\\
		&NN-TRPCA&\textbf{30.71}&\textbf{0.958}&\textbf{0.966}\\
		\midrule[0.4pt]
		\multirow{6}*{Stone}&SNN-TRPCA&26.41&0.733&0.793\\		
		&KBR-TRPCA&26.78&0.797&0.881\\
		&TNN-TRPCA&27.27&0.802&0.879\\
		&TPSCPSF&28.42&0.828&0.885\\
		&N-TRPCA&29.18&0.824&0.892\\
		&NN-TRPCA&\textbf{34.90}&\textbf{0.950}&\textbf{0.961}\\
		\midrule[0.4pt]
		\multirow{6}*{Boat}&SNN-TRPCA&23.67&0.831&0.821\\		
		&KBR-TRPCA&26.45&0.885&0.897\\
		&TNN-TRPCA&25.65&0.881&0.878\\
		&TPSCPSF&29.10&0.913&0.926\\
		&N-TRPCA&28.01&0.899&0.910\\
		&NN-TRPCA&\textbf{32.21}&\textbf{0.945}&\textbf{0.960}\\
		\midrule[0.4pt]
		\multirow{6}*{Pyramid}&SNN-TRPCA&26.82&0.754&0.738\\		
		&KBR-TRPCA&28.73&0.851&0.883\\
		&TNN-TRPCA&28.67&0.845&0.843\\
		&TPSCPSF&30.83&0.861&0.871\\
		&N-TRPCA&29.73&0.839&0.862\\
		&NN-TRPCA&\textbf{33.35}&\textbf{0.924}&\textbf{0.928}\\
		\bottomrule[0.4pt]
	\end{tabular}	
\end{table}

The parameters of all experiments are set as follows. $[\lambda_1,\lambda_2,\lambda_3]$ in SNN-TRPCA is empirically set to $[15,15,1.5]$ for color image restoration. This setting can enable SNN-TRPCA to perform well in most image cases. For video restoration, $[\lambda_1,\lambda_2,\lambda_3]$ are set differently since the three videos have different correlations along each mode. Specifically, $[\lambda_1,\lambda_2,\lambda_3]$ are set to $[12,12,17]$, $[13,13,20]$, and $[12,12,16]$ for 'Hall $\&$ Monitor', 'Candela$\_$m1.10', and 'CAVIAR1', respectively. For TNN-TRPCA and KBR-TRPCA, we follow the default parameters setting suggested by their authors. The parameter $\kappa$ in TPSCPSF is set to $0.9$ for color image restoration and $0.6$ for video restoration. For N-TRPCA, we empirically set $\mu_0=1e-3$, $\mu_{max}=1e10$, $\rho=1.1$, $\epsilon=1e-5$, and $\theta=2$, respectively. And as used in \cite{re5}, our parameter $\lambda$ is set to $1/\sqrt{\max(n_1,n_2)n_3}$. For NN-TRPCA, the patch size $p$ and the number of patches $m$ are empirically set to 10 and 100.

\subsection{Results on Color Image Restoration}\label{sec5.2}

Table~\ref{tab_result1} displays the average quantitative results of competitive methods on test images with different noise rates $(10\%,20\%,30\%)$. It can be observed that KBR-TRPCA outperforms SNN-TRPCA evidently. The main reason is that, KBR-TRPCA combines the advantage of CP rank and Tucker rank. In addition, TNN-TRPCA is competitive with KBR-TRPCA and better than SNN-TRPCA. This is practically reasonable because TNN-TRPCA successfully captures the multidimensional structural information in tensors equipped with the tensor average rank. TPSCPSF and N-TRPCA are superior to SNN-TRPCA, TNN-TRPCA and KBR-TRPCA in all evaluation indices. This can be attributed to the factors: 1) TPSCPSF makes full use of side information and features; 2) N-TRPCA well preserves the siginificant information in tensors by shrinking the tensor singular values differently. More importantly, NN-TRPCA has yielded the best performance. This is due to the fact that NN-TRPCA takes full use of the structural redundancy in color images by introducing the nonlocal prior.

For more intuitive comparison, the restoration results of different methods on eight example images are shown in Fig. \ref{Fig_result1}. One can observe that SNN-TRPCA produces serious block artifacts in all scenarios. This is because directly unfolding tensors along each mode will lose structural information in tensor data. The images restored by N-TRPCA also contain some artifacts like by TNN-TRPCA. Fortunately, by introducing the  nonlocal self-similarity, NN-TRPCA removes the artifacts and yields the best visual effect. It is worth noting that, NN-TRPCA can well restore the detail information of complex images, e.g., texture of starfish, water-drops on flowers, patterns on girl's clothes, and the edge of pyramid. Futhermore, the quantitative results on these example images are recorded in Table~\ref{tab_result2}, from which we makes the following observations. Although N-TRPCA produces artifacts like TNN-TRPCA, N-TRPCA is far superior to TNN-TRPCA in terms of all the evaluation indices, which confirms the effectiveness of the tensor adjustable logarithmic norm to retain the important information in color images. Besides, our NN-TRPCA obtains the best evaluation indices over all competing methods by utilizing the nonlocal redundancy of natural color images. Especially for images with complex structures, such as `Starfish', `Flower', and `Girl', NN-TRPCA still has a significant improvement compared with N-TRPCA. This is due to the fact that the nonlocal self-similarity is abundant in these complex images. By grouping nonlocal similar patches, the group tensor is strongly low-rank, which offers contribution for better restoration. In a word, the proposed NN-TRPCA method delivers the best recovery performance on both visual results and evaluation indices.

\begin{table}[!t]
	\centering
	\caption{Quantitative Performance Comparison of Different Algorithms on Three Videos. }
	\label{tab_result3}
	\begin{tabular}{m{2.0 cm}<{\centering} m{2.3 cm}<{\centering} m{0.7 cm}<{\centering} m{0.7 cm}<{\centering} m{0.85 cm}<{\centering}}
		\toprule
		Video&Method&PSNR&SSIM&FSIM\\
		\midrule[0.4pt]
		\multirow{5}*{Hall $\&$ Monitor}&SNN-TRPCA&26.08&0.911&0.941\\
		&KBR-TRPCA& 26.13&0.923&0.955\\
		&TNN-TRPCA&31.43&0.973&0.988\\
		&TPSCPSF&30.81&0.951&0.974\\
		&N-TRPCA&34.28&0.981&0.992\\
		&NN-TRPCA&\textbf{37.76}&\textbf{0.991}&\textbf{0.998}\\
		\midrule[0.4pt]
		\multirow{6}*{Candela$\_$m1.10}&SNN-TRPCA&25.65& 0.848& 0.938\\
		&KBR-TRPCA&26.78& 0.910& 0.967\\
		&TNN-TRPCA&30.96& 0.969& 0.993\\
		&TPSCPSF&35.03&0.976&0.988\\
		&N-TRPCA&34.28& 0.983&0.996\\
		&NN-TRPCA&\textbf{38.74}& \textbf{0.993}& \textbf{0.997}\\
		\midrule[0.4pt]
		\multirow{5}*{CAVIAR1}&SNN-TRPCA&29.95 &0.941 &0.940\\
		&KBR-TRPCA&31.02 &0.951 &0.946\\
		&TNN-TRPCA&34.99 &0.983 &0.970\\
		&TPSCPSF&35.49&0.975&0.976\\
		&N-TRPCA&40.75 &0.995 &0.988\\
		&NN-TRPCA&\textbf{42.25} &\textbf{0.996}& \textbf{0.992}\\
		\bottomrule[0.4pt]
	\end{tabular}
\end{table}

\subsection{Results on Gray Video Restoration}\label{sec5.3}

\begin{figure*}[!t]
	\centering
	\subfloat{
		\begin{minipage}[ht]{0.11\linewidth}
			\centering
			\includegraphics[width=2.15 cm]{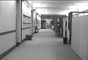}\\
			\vspace{0.07cm}
			\includegraphics[width=2.15 cm]{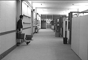}\\
			\vspace{0.07cm}
			\includegraphics[width=2.15 cm]{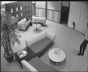}\\
			\vspace{0.07cm}
			\includegraphics[width=2.15 cm]{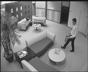}\\
			\vspace{0.07cm}
			\includegraphics[width=2.15 cm]{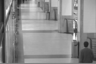}\\
			\vspace{0.07cm}
			\includegraphics[width=2.15 cm]{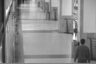}\\
			\centerline{\small Original}
		\end{minipage}
	}	  
	\subfloat{
		\begin{minipage}[ht]{0.11\linewidth}
			\centering
			\includegraphics[width=2.15 cm]{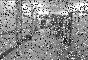}\\
			\vspace{0.07cm}
			\includegraphics[width=2.15 cm]{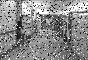}\\
			\vspace{0.07cm}
			\includegraphics[width=2.15 cm]{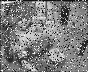}\\
			\vspace{0.07cm}
			\includegraphics[width=2.15 cm]{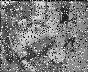}\\
			\vspace{0.07cm}
			\includegraphics[width=2.15 cm]{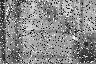}\\
			\vspace{0.07cm}
			\includegraphics[width=2.15 cm]{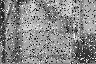}\\
			\centerline{\small Corrupted}
		\end{minipage}
	} 
	\subfloat{
		\begin{minipage}[ht]{0.11\linewidth}
			\centering
			\includegraphics[width=2.15 cm]{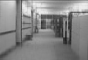}\\
			\vspace{0.07cm}
			\includegraphics[width=2.15 cm]{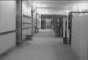}\\
			\vspace{0.07cm}
			\includegraphics[width=2.15 cm]{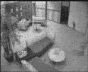}\\
			\vspace{0.07cm}
			\includegraphics[width=2.15 cm]{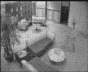}\\
			\vspace{0.07cm}
			\includegraphics[width=2.15 cm]{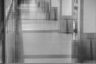}\\
			\vspace{0.07cm}
			\includegraphics[width=2.15 cm]{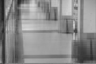}\\
			\centerline{\small SNN-TRPCA}
		\end{minipage}
	} 	
	\subfloat{
		\begin{minipage}[ht]{0.11\linewidth}
			\centering
			\includegraphics[width=2.15 cm]{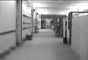}\\
			\vspace{0.07cm}
			\includegraphics[width=2.15 cm]{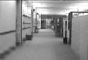}\\
			\vspace{0.07cm}
			\includegraphics[width=2.15 cm]{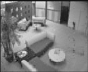}\\
			\vspace{0.07cm}
			\includegraphics[width=2.15 cm]{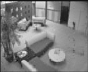}\\
			\vspace{0.07cm}
			\includegraphics[width=2.15 cm]{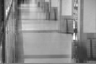}\\
			\vspace{0.07cm}
			\includegraphics[width=2.15 cm]{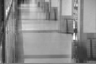}\\
			\centerline{\small KBR-TRPCA}
		\end{minipage}
	}
	\subfloat{
		\begin{minipage}[ht]{0.11\linewidth}
			\centering
			\includegraphics[width=2.15 cm]{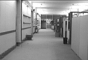}\\
			\vspace{0.07cm}
			\includegraphics[width=2.15 cm]{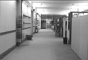}\\
			\vspace{0.07cm}
			\includegraphics[width=2.15 cm]{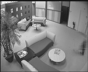}\\
			\vspace{0.07cm}
			\includegraphics[width=2.15 cm]{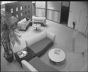}\\
			\vspace{0.07cm}
			\includegraphics[width=2.15 cm]{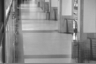}\\
			\vspace{0.07cm}
			\includegraphics[width=2.15 cm]{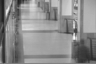}\\
			\centerline{\small TNN-TRPCA}
		\end{minipage}
	} 
	\subfloat{
		\begin{minipage}[ht]{0.11\linewidth}
			\centering
			\includegraphics[width=2.15 cm]{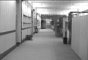}\\
			\vspace{0.07cm}
			\includegraphics[width=2.15 cm]{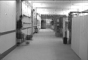}\\
			\vspace{0.07cm}
			\includegraphics[width=2.15 cm]{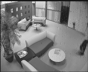}\\
			\vspace{0.07cm}
			\includegraphics[width=2.15 cm]{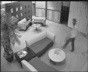}\\
			\vspace{0.07cm}
			\includegraphics[width=2.15 cm]{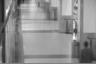}\\
			\vspace{0.07cm}
			\includegraphics[width=2.15 cm]{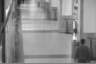}\\
			\centerline{\small TPSCPSF}
		\end{minipage}
	}  
	\subfloat{
		\begin{minipage}[ht]{0.11\linewidth}
			\centering
			\includegraphics[width=2.15 cm]{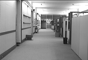}\\
			\vspace{0.07cm}
			\includegraphics[width=2.15 cm]{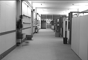}\\
			\vspace{0.07cm}
			\includegraphics[width=2.15 cm]{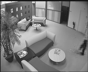}\\
			\vspace{0.07cm}
			\includegraphics[width=2.15 cm]{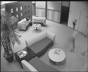}\\
			\vspace{0.07cm}
			\includegraphics[width=2.15 cm]{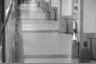}\\
			\vspace{0.07cm}
			\includegraphics[width=2.15 cm]{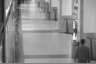}\\
			\centerline{\small N-TRPCA}
		\end{minipage}
	} 
	\subfloat{
		\begin{minipage}[ht]{0.11\linewidth}
			\centering
			\includegraphics[width=2.15 cm]{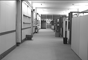}\\
			\vspace{0.07cm}
			\includegraphics[width=2.15 cm]{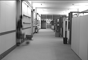}\\
			\vspace{0.07cm}
			\includegraphics[width=2.15 cm]{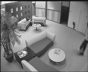}\\
			\vspace{0.07cm}
			\includegraphics[width=2.15 cm]{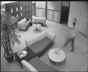}\\
			\vspace{0.07cm}
			\includegraphics[width=2.15 cm]{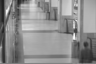}\\
			\vspace{0.07cm}
			\includegraphics[width=2.15 cm]{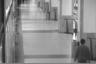}\\
			\centerline{\small  NN-TRPCA}
		\end{minipage}
	} 
	\centering
	\caption{Several frames of restoration results on three videos. From top to bottom: `Hall $\&$ Monitor', `Candela$\_$m1.10', and `CAVIAR1'. }
	\label{Fig_result2}
\end{figure*}

Fig. \ref{Fig_result2} lists several frames of restoration results of different algorithms on test videos with noise rate $30\%$. These videos capture a walking human but in three different scenes. From the restoration results, we can observe that, TPSCPSF well restores the walking people. This is because TPSCPSF fully utilizes two kinds of side information. Our N-TRPCA more clearly restores the important structure information of videos compared with TNN-TRPCA. The reason lies in that the tensor adjustable logarithmic norm used in N-TRPCA is capable to preserve the important information in videos by shrinking the large singular values less and the small ones more. It can be further found that, although both N-TRPCA and NN-TRPCA can recover the main structure of walking humans, NN-TRPCA restores the contour of the walking man more accurately. This indicates that the introduction of nonlocal self-similarity enables NN-TRPCA to recover more detail information in videos.

Meanwhile, as shown in Table~\ref{tab_result3}, the proposed N-TRPCA and NN-TRPCA have yielded very competitive scores of evaluation indices. More specifically, our N-TRPCA can significantly outperform other competitive methods, e.g., N-TRPCA achieves 3.98 dB gain on PSNR beyond TNN-TRPCA on average of the three videos. This is due to the fact that N-TRPCA treats the tensor singular values differently. Furthermore, by integrating the nonlocal self-similarity, the performance of NN-TRPCA is improved, i.e., averagely by 3.15 dB gain on PSNR, 0.007 on SSIM and 0.004 on FSIM. All these scores validate the superiority and effectiveness of our NN-TRPCA method.

\subsection{Impacts of Parameters}\label{sec5.4}

\begin{figure*}[!h]
	\centering
	\subfloat{
		\begin{minipage}[ht]{0.235\linewidth}
			\centering
			\includegraphics[width=4.6 cm]{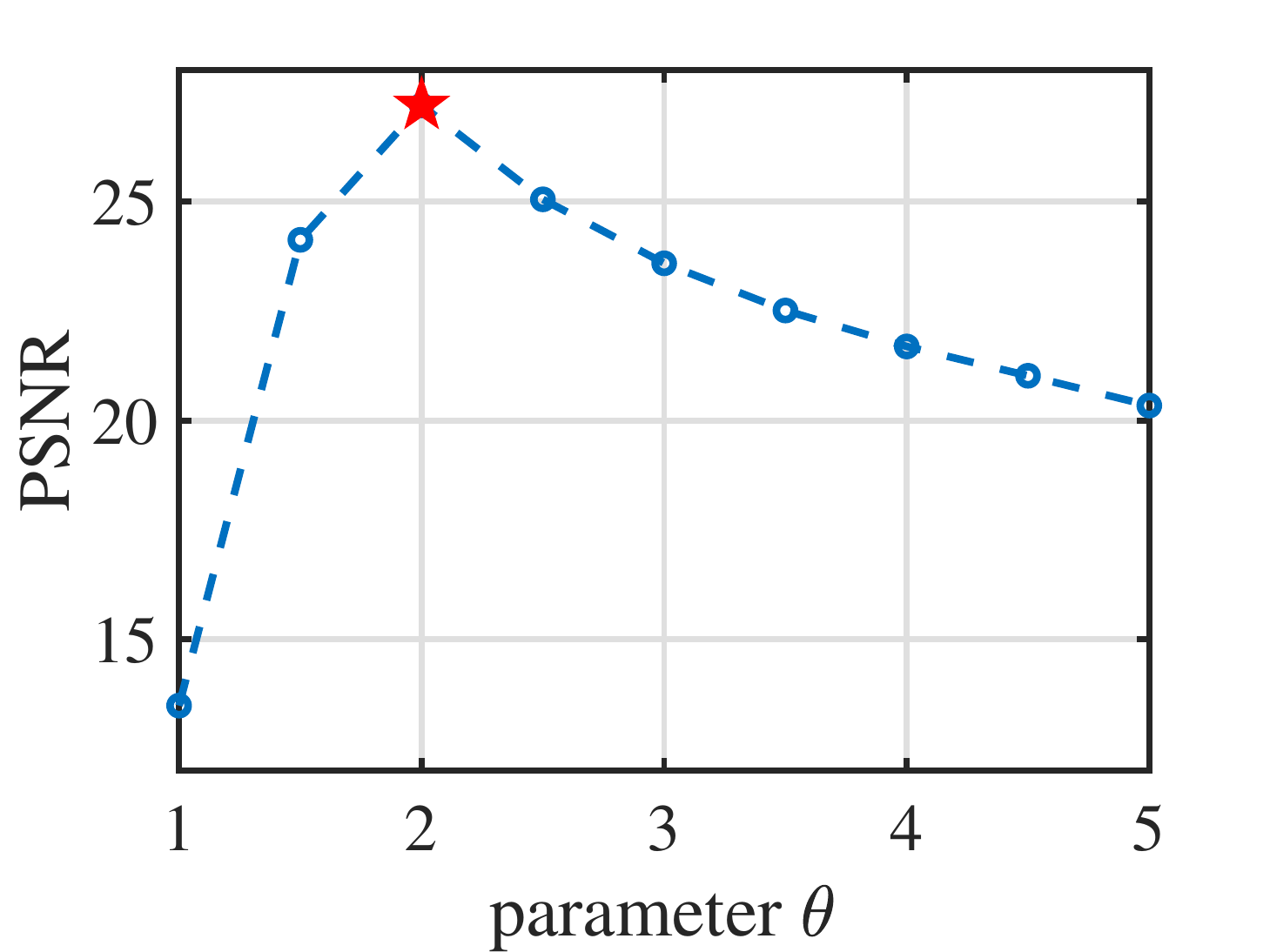}
		\end{minipage}
	}	  
	\subfloat{
		\begin{minipage}[ht]{0.235\linewidth}
			\centering
			\includegraphics[width=4.6 cm]{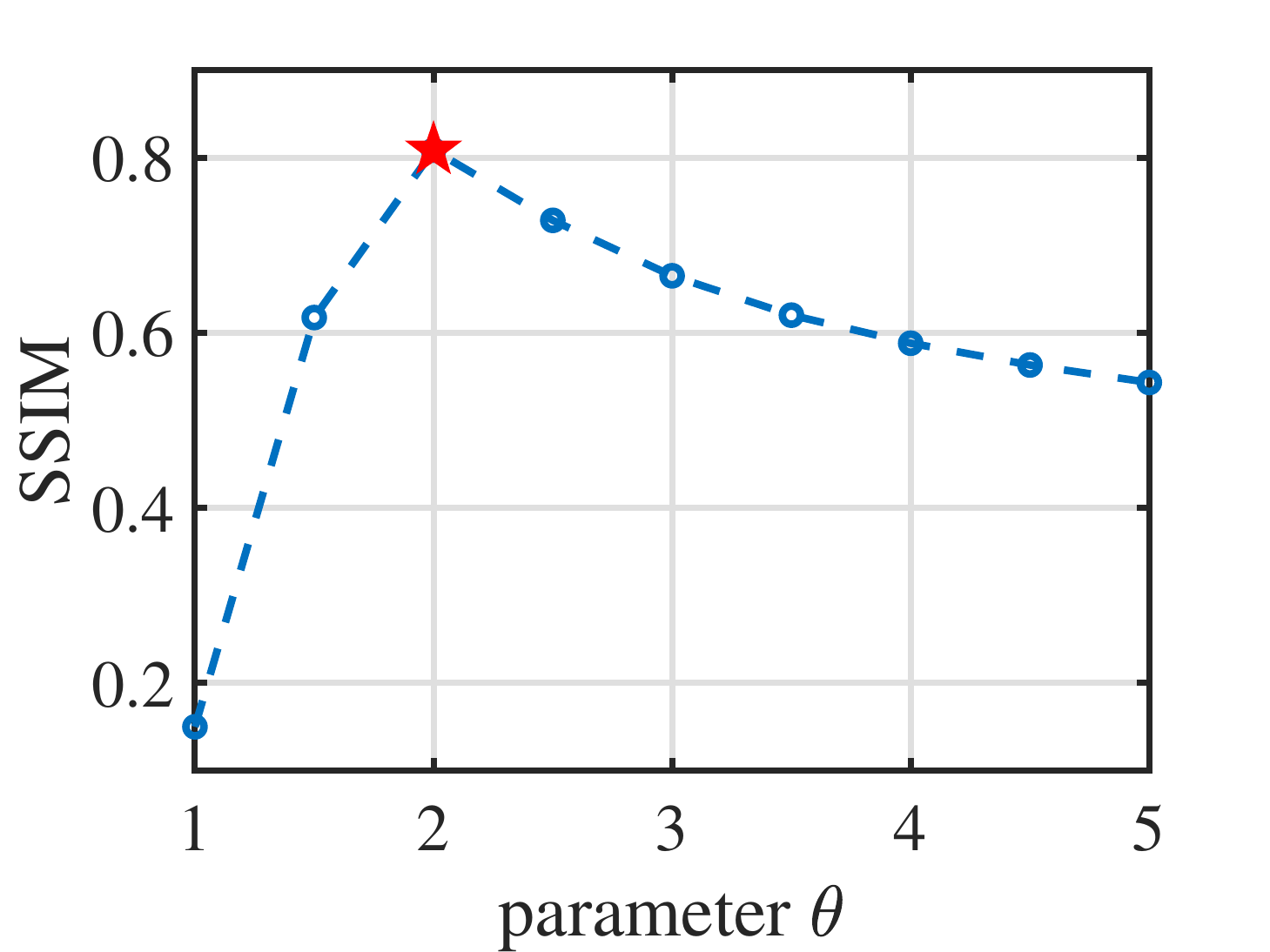}
		\end{minipage}
	} 
	\subfloat{
		\begin{minipage}[ht]{0.235\linewidth}
			\centering
			\includegraphics[width=4.6 cm]{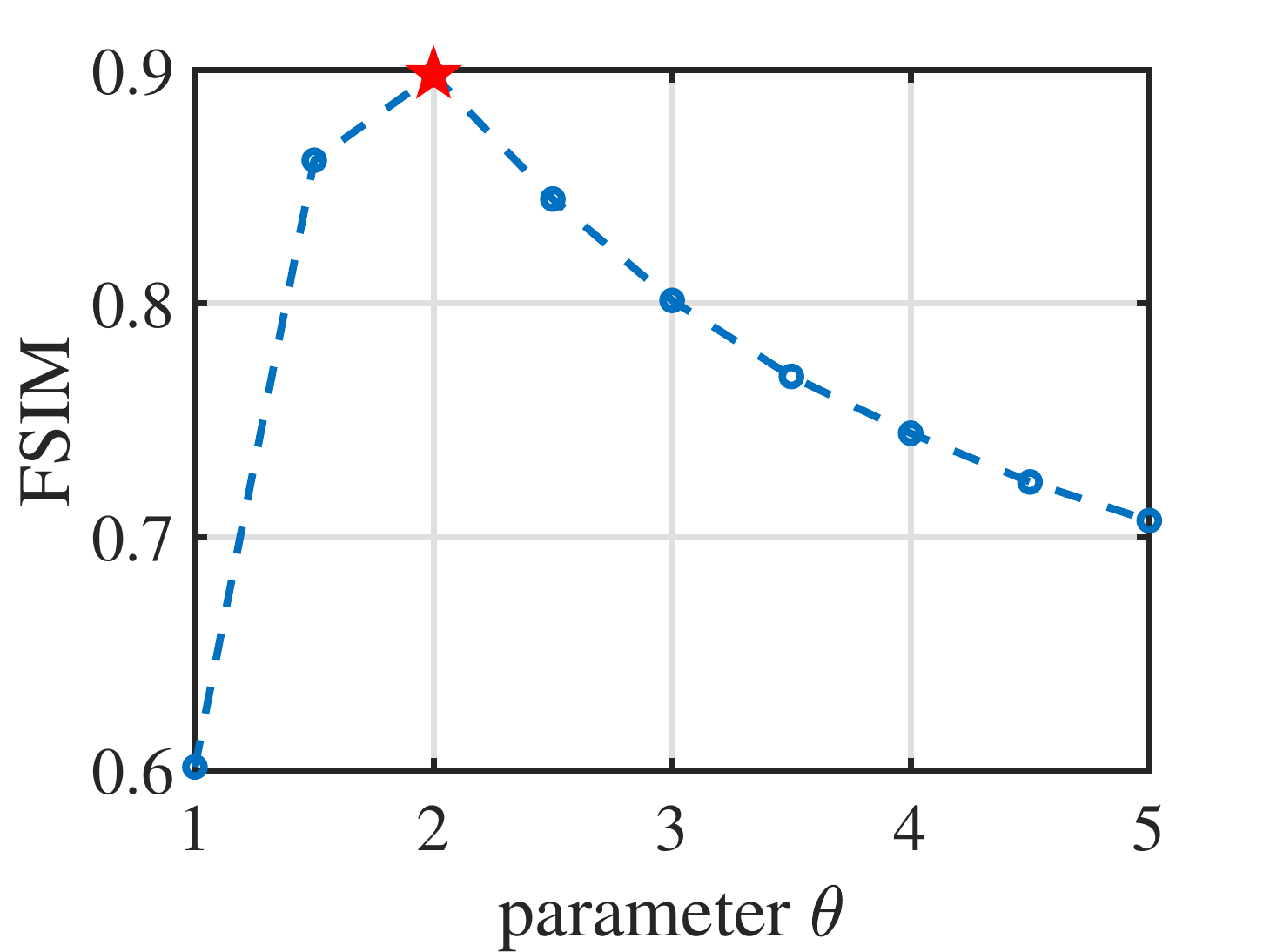}
		\end{minipage}
	}  
	\centering
	\caption{Quantitative Comparison of the Effect of Parameter $\theta$ on N-TRPCA.}
	\label{Fig_result3}
\end{figure*}

In NN-TRPCA, there are three algorithmic parameters, i.e., adjustable parameter $\theta$, patch size $p$, and the number of patches in each group tensor $m$. The parameter $\theta$ controls the shrinkage level to tensor singular values. The larger $\theta$ indicates more shrinkage. To analyze the impact of nonconvex parameter $\theta$, we fix other parameters and perform N-TRPCA on the test images with different values $\theta=1,1.5,2,\ldots,5$. The average change curves of four PQIs with varying $\theta$ are plot in Fig. \ref{Fig_result3}. From them, we can see that the best restoration results are obtained at $\theta=2$. Thus, the parameter $\theta$ is empirically set to $2$.

\begin{figure*}[!h]
	\centering
	\subfloat{
		\begin{minipage}[ht]{0.235\linewidth}
			\centering
			\includegraphics[width=4.6 cm]{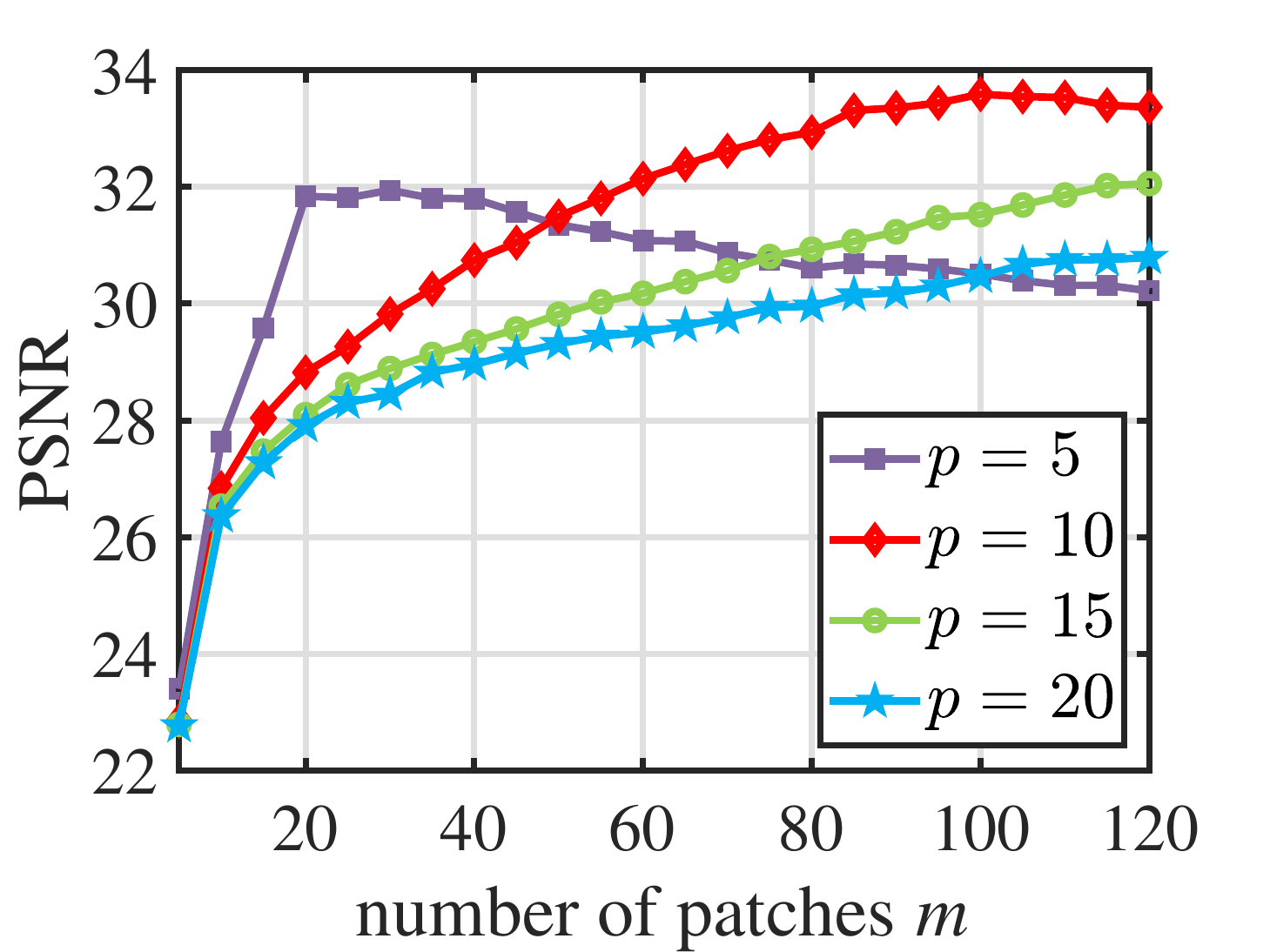}
		\end{minipage}
	}	  
	\subfloat{
		\begin{minipage}[ht]{0.235\linewidth}
			\centering
			\includegraphics[width=4.6 cm]{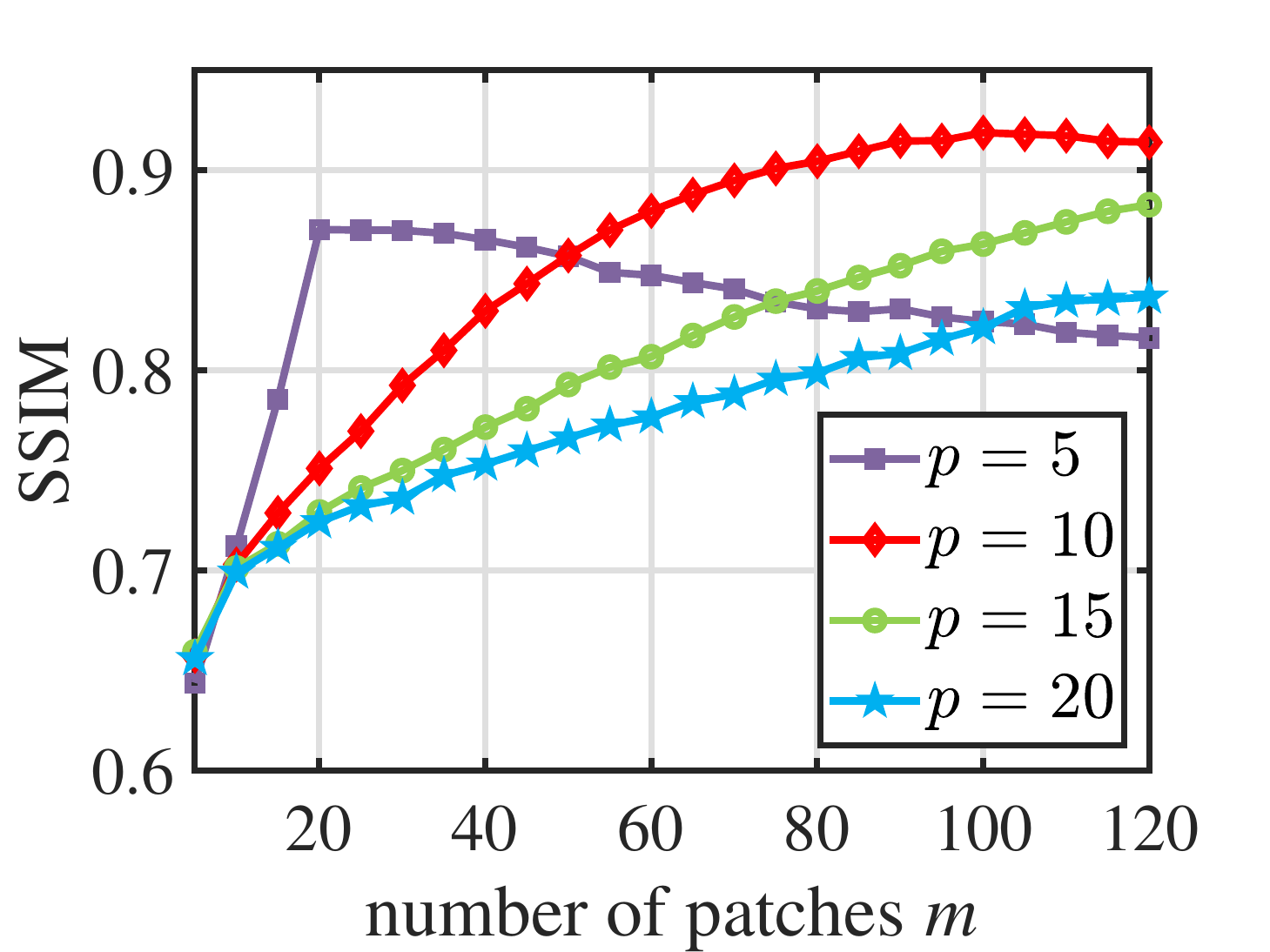}
		\end{minipage}
	} 
	\subfloat{
		\begin{minipage}[ht]{0.235\linewidth}
			\centering
			\includegraphics[width=4.6 cm]{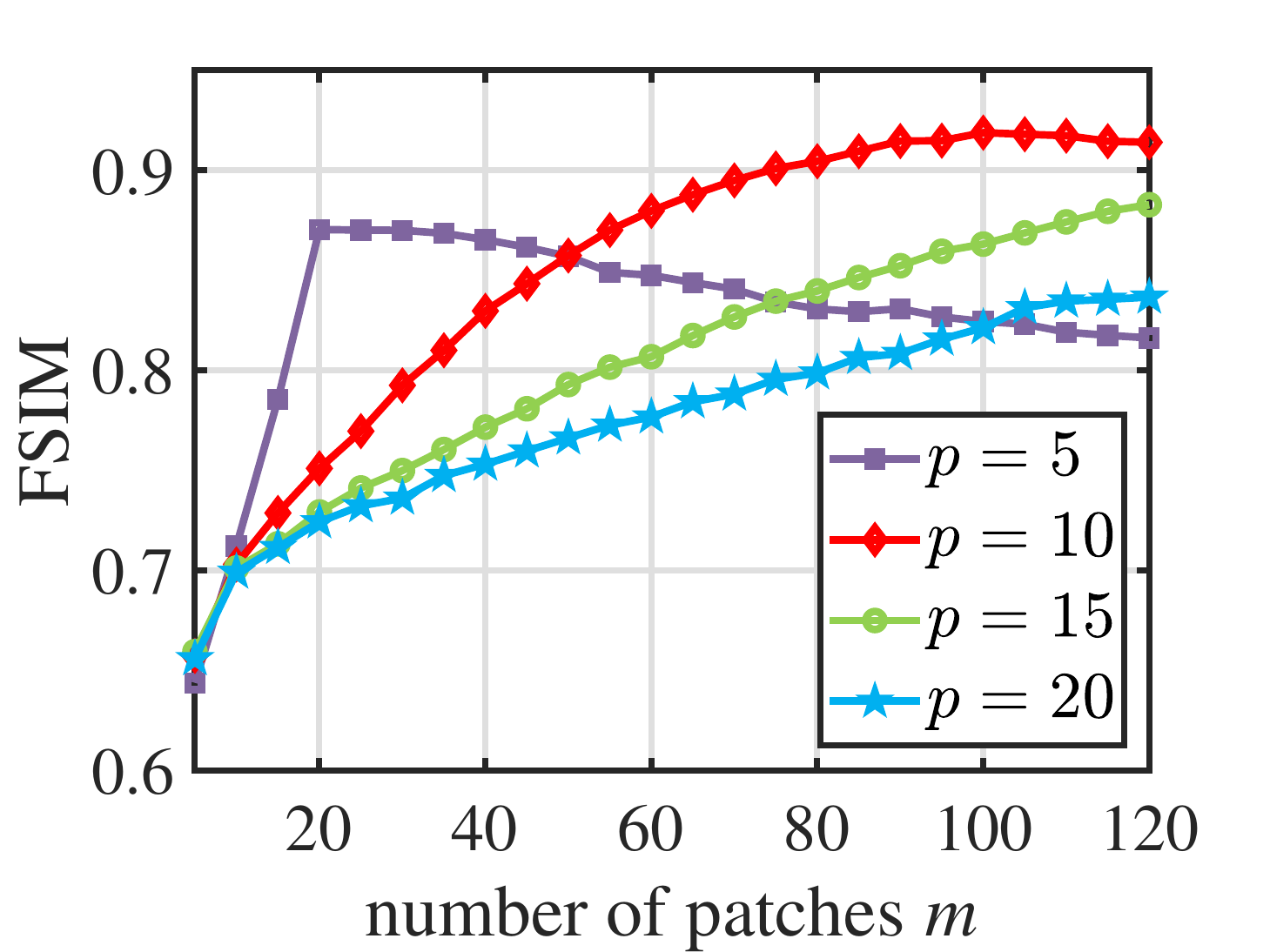}
		\end{minipage}
	} 
	\caption{Quantitative Comparison of the Effect of Parameters $p$ and $m$ on NN-TRPCA.}
	\label{parameters}
\end{figure*}

The patch size $p$ and the number of patches in each group tensor $m$ are two important parameters to capture the nonlocal redundancy in images. When $p$ is too small, it is hard to preserve the local structure of patches. When $p$ is too large, there are too many details in each patch, resulting in the reduction of the similarity of patches in one group. Besides, a too small $m$ could divide the closely similar patches into different groups and a too large $m$ could classify other dissimilar patches into one similar group. Therefore, a too small $m$ or a too large $m$ will affect the accuracy of patch grouping, leading to a degraded restoration performance. To study the influence of parameters $p$ and $m$, we run our NN-TRPCA algorithm on the test image 'Elephant' with different values of $p$ and $m$. Fig. \ref{parameters} displays the average change curves of four PQIs with varying value pairs $(p,m)$. It can be found that NN-TRPCA with parameter pair $(p,m)=(10,100)$ can yield the best restoration performance. Hence, in our experiments, $p$ and $m$ are fixed to 10 and 100, respectively.

\subsection{Computational Cost}\label{sec5.5}

We also compare the computational efficiency of the proposed algorithms to those of three state-of-the-art TRPCA algorithms. Table \ref{Running_time} summarizes the average running times of different algorithms on the aforementioned tasks. From it, one can observe that the running time of our N-TRPCA is slight higher than that of TNN-TRPCA. This is because TNN-TRPCA algorithm simply uses a fixed threshold to all the singular values shrinkage while our N-TRPCA adaptively calculates the threshold for each singular value to shrink the small singular values more and the large ones less. Note that the computational cost of our NN-TRPCA is high expensive. The main computational cost of NN-TRPCA is the calculation of t-SVD for each nonlocal group tensor, and the running time of the patch grouping stage as well as the aggregation stage can be negligible. In fact, since the t-SVD of each group tensor is calculated independently, a direct acceleration way is parallel GPU implementation for the t-SVDs. Besides, as used in \cite{re54} and \cite{re55}, the Lanczos bidiagonalization process can be employed to accelerate the calculation of SVD by only considering the few dominant singular values.

\section{CONCLUSION AND FUTURE WORK}\label{sec6}

TRPCA, aiming to restore a low-rank tensor from it corrupted version, has attracted considerable interest in the fields of image and video processing. In this work, we firstly build the N-TRPCA based on the tensor logarithmic norm which can better retain the significant information of visual data. Furthermore, to fully utilize the structural redundancy in visual data, we propose the NN-TRPCA by introducing the nonlocal self-similarity. This model can well recover the edges and textures in images and videos. Meanwhile, an efficient algorithm based on ADMM is designed to solve the proposed models. Extensive experimental results confirm the effectiveness of our models compared with recent state-of-the-art TRPCA models.

For future work, there are two possible expansion directions. Firstly, this paper focuses on three-dimensional visual data due to the fact that the t-SVD and tensor average rank are defined on three-way tensors. Therefore, one can generalize the t-SVD and tensor average rank for higher-dimensional tensors, and further propose a higher-order version of the NN-TRPCA. Secondly, in order to obtain lower-rank tensors, some studies have attempted to replace FFT used in t-SVD with other invertible linear transform, such as framelet transform \cite{re72}, discrete cosine transform \cite{re73}. Inspired by these works, we will attempt to find a more suitable transform to capture the low-rankness of visual data, so as to achieve better performance of visual data restoration.

\begin{table*}[!t]
	\centering
	\caption{Average Running Times (in Seconds) of Different Algorithms.}
	\label{Running_time}
	\begin{tabular}{m{2.6 cm}<{\centering} m{2.11 cm}<{\centering} m{2.17 cm}<{\centering} m{2.17 cm}<{\centering} m{2 cm}<{\centering} m{2 cm}<{\centering} m{2 cm}<{\centering}}
		\toprule
		Task &SNN-TRPCA&KBR-TRPCA&TNN-TRPCA&TPSCPSF&N-TRPCA&NN-TRPCA\\
		\midrule[0.4pt]
		Image restoraton&9.98&97.42&10.62&565.72&13.63&2679.98\\
		Video restoraton&2.23&2.87&2.40&489.94&4.04&1431.27\\
		\bottomrule[0.4pt]
	\end{tabular}
\end{table*}

\section*{Acknowledgments}
This work was supported by the National Natural Science Foundation of China under Grant 61873145.




%\bibliographystyle{elsarticle-num}
%\bibliography{re}

\end{document}